\documentclass[runningheads]{llncs}
\usepackage{graphicx}
\usepackage{tikz}
\usepackage{comment}
\usepackage{color}
\usepackage[linesnumbered, ruled,vlined]{algorithm2e}
\usepackage{etoolbox}
\usepackage{subcaption}
\usepackage{enumitem}
\usepackage{xscommand}
\usepackage{algorithmicx}
\usepackage[linesnumbered, ruled,vlined]{algorithm2e}
\usepackage{multirow}
\usepackage{booktabs}
\usepackage{wrapfig}
\usepackage{comment}
\usepackage{color}
\usepackage[accsupp]{axessibility}  % Improves PDF readability for those with disabilities.
\usepackage[toc,page]{appendix}
\usepackage{bm}
\usepackage{graphicx}
\makeatletter
\patchcmd{\@algocf@start}% <cmd>
  {-1.5em}% <search>
  {0pt}% <replace>
  {}{}% <success><failure>
  
  \patchcmd\algocf@Vline{\vrule}{\vrule \kern-0.4pt}{}{}
\patchcmd\algocf@Vsline{\vrule}{\vrule \kern-0.4pt}{}{}
\makeatother

\begin{document}
% \linenumbers
\pagestyle{headings}
\mainmatter
\def\ECCVSubNumber{6276}  % Insert your submission number here

\title{Triangle Attack: A Query-efficient Decision-based Adversarial Attack} % Replace with your title

\begin{comment}
\titlerunning{ECCV-22 submission ID \ECCVSubNumber} 
\authorrunning{ECCV-22 submission ID \ECCVSubNumber} 
\author{Anonymous ECCV submission}
\institute{Paper ID \ECCVSubNumber}
\end{comment}
%******************

% \begin{comment}
\titlerunning{Triangle Attack: A Query-efficient Decision-based Adversarial Attack}
\author{Xiaosen Wang\inst{1,2} \and
Zeliang Zhang\inst{1} \and Kangheng Tong\inst{1} \and Dihong Gong\inst{2} \and Kun He\inst{1}\thanks{Corresponding authors} \and Zhifeng Li\inst{2}\textsuperscript{$\star$} \and Wei Liu\inst{2}\textsuperscript{$\star$}}
\authorrunning{X. Wang et al.}
\institute{School of Computer Science and Technology, HUST \and
Data Platform, Tencent}
% \end{comment}
%******************
\maketitle

\begin{abstract}
Decision-based attack poses a severe threat to real-world applications since it regards the target model as a black box and only accesses the hard prediction label. Great efforts have been made recently to decrease the number of queries; however, existing decision-based attacks still require thousands of queries in order to generate good quality adversarial examples. In this work, we find that a benign sample, the current and the next adversarial examples can naturally construct a triangle in a subspace for any iterative attacks. Based on the law of sines, we propose a novel Triangle Attack (\name)  to optimize the perturbation by utilizing the geometric information that the longer side is always opposite the larger angle in any triangle. However, directly applying such information on the input image is ineffective because it cannot thoroughly explore the neighborhood of the input sample in the high dimensional space. To address this issue, \name optimizes the perturbation in the low frequency space for effective dimensionality reduction owing to the generality of such geometric property. Extensive evaluations on ImageNet dataset show that \name achieves a much higher attack success rate within 1,000 queries and needs a much less number of queries to achieve the same attack success rate under various perturbation budgets than existing decision-based attacks. With such high efficiency, we further validate the applicability of \name on real-world API, \ie, \app.
\end{abstract}

\section{Introduction}

Despite the unprecedented progress of Deep Neural Networks (DNNs)~\cite{krizhevsky2012imagenet,he2016deep,huang2017densely}, the vulnerability to adversarial examples~\cite{szegedy2013intriguing} poses serious threats to security-sensitive applications, \eg, face recognition~\cite{sharif2016accessorize,tang2004video,gong2013multi,li2014common,wen2016discriminative,wang2018cosface,deng2019mutual,qiu2021end2end,yang2021larnet}, autonomous driving~\cite{chen2015deepdriving,eykholt2018robust,bojarski2016end,xu2017end,sallab2017deep}, \etc. To securely deploy DNNs in various real-world applications, it is necessary to conduct an in-depth analysis on the intrinsic properties of adversarial examples, which has inspired numerous researches on adversarial attacks~\cite{moosavi2016deepfool,carlini2017towards,athalye2018obfuscated,cheng2018query,chen2020hopskipjumpattack,dong2018boosting,brendel2017decision,wang2021admix} and defenses~\cite{madry2017towards,guo2017countering,zhang2019theoretically,wong2020fast,wu2021attacking,wang2021Natural}. Existing attacks can be split into two categories: \textit{white-box} attack has full knowledge of the target model (often leveraging the gradient)~\cite{goodfellow2014explaining,carlini2017towards,madry2017towards,dong2018boosting} while \textit{black-box} attack can only access the model output, which is more applicable in real-world scenarios. The black-box attack can be implemented in different ways. \textit{Transfer-based} attack~\cite{liu2016delving,dong2018boosting,xie2019improving,wei2018transferable} adopts the adversaries generated on the substitute model to fool the target model directly. \textit{Score-based} attack~\cite{chen2017zoo,ilyas2018black,al2019sign,liang2022parallel} assumes that the attacker can access the output logits while \textit{decision-based} (\aka hard label) attack~\cite{brendel2017decision,cheng2018query,chen2020boosting,li2020qeba,maho2021surfree} only has access to the prediction (top-1) label.

Among the black-box attacks, decision-based attack is more challenging and practical due to the minimum information requirement for attack. The number of queries on target model often plays a significant role in decision-based attack, since the access to a victim model is usually restricted in practice. Though recent works manage to reduce the total number of queries from millions to thousands of requests~\cite{brendel2017decision,li2020qeba,rahmati2020geoda}, it is still insufficient for most practical applications~\cite{maho2021surfree}. 

\begin{wrapfigure}{R}{0.465\textwidth}
    \centering
    \vspace{-2.5em}
    \includegraphics[width=0.9\linewidth]{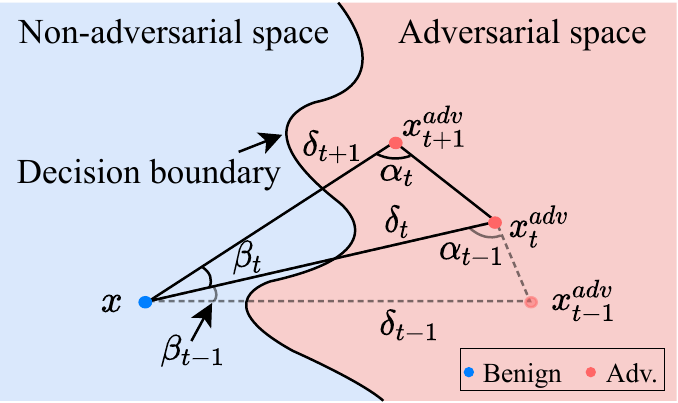}
    \caption{Illustration of the candidate triangle at an arbitrary iteration of \name. At the $t$-th iteration, \name constructs a triangle with the learned angle $\alpha_t$ which satisfies $\beta_t + 2\alpha_t > \pi$ in the sampled subspace to find a new adversarial example $x_{t+1}^{adv}$ and update $\alpha_t$ accordingly. Note that different from existing decision-based attacks~\cite{brendel2017decision,rahmati2020geoda,maho2021surfree}, \name does not restrict $x_t^{adv}$ on the decision boundary but minimizes the perturbation in the low frequency space using the geometric property; making \name itself query-efficient}
    \label{fig:TA}
    \vspace{-3em}
\end{wrapfigure}

Existing decision-based attacks~\cite{brendel2017decision,li2020qeba,rahmati2020geoda,maho2021surfree} first generate a large adversarial perturbation and then minimize the perturbation while keeping  adversarial property by various optimization methods. As shown in Fig.~\ref{fig:TA}, we find that at the $t$-th iteration, the benign sample $x$, current adversarial example $x_t^{adv}$, and next adversarial example $x_{t+1}^{adv}$ can naturally construct a triangle for any iterative attacks. According to the law of sines, the adversarial example $x_{t+1}^{adv}$ at the $(t+1)$-th iteration should satisfy  $\beta_t + 2\alpha_t > \pi$ to 
guarantee that the perturbation decreases, \ie, $\delta_{t+1}=\|x_{t+1}^{adv}-x\|_p<\delta_t=\|x_t^{adv}-x\|_p$ (when $\beta_t + 2\cdot \alpha_t = \pi$, it would be an isosceles triangle, \ie, $\delta_{t+1}=\delta_{t}$).

Based on the above geometric property, we propose a novel and query-efficient decision-based attack, called Triangle Attack (TA). Specifically, at $t$-th iteration, we randomly select a directional line across the benign sample $x$ to determine a 2-D subspace, in which we iteratively construct the triangle based on the current adversarial example $x_t^{adv}$, benign sample $x$, learned angle $\alpha_t$, and searched angle $\beta_t$ until the third vertex of the constructed triangle is adversarial. Using the geometric information, we can conduct \name in the low frequency space generated by Discrete Cosine Transform (DCT)~\cite{ahmed1974discrete} for effective dimensionality reduction to improve the efficiency. And we further update $\alpha_t$ to adapt to the perturbation optimization for each constructed triangle. Different from most existing decision-based attacks, there is no need to restrict $x_t^{adv}$ on the decision boundary or estimate the gradient at each iteration, making \name query-efficient.

Our main contributions are summarized as follows:
\begin{itemize}
    \setlength{\itemsep}{0pt}
    \setlength{\parsep}{0.1pt}
    \setlength{\parskip}{0pt}
    \vspace{-0.6em}
    \item To the best of our knowledge, it is the first work that directly optimizes the perturbation in frequency space via geometric information without restricting the adversary on decision boundary, leading to high query efficiency.
    \item Extensive evaluations on ImageNet dataset show that \name exhibits a much higher attack success rate within 1,000 queries and needs a much less number of queries to achieve the same attack success rate with the same perturbation budget on five models than existing SOTA attacks~\cite{cheng2018query,cheng2019sign,chen2020hopskipjumpattack,li2020qeba,rahmati2020geoda,maho2021surfree}.
    \item \name generates more adversarial examples with imperceptible perturbations on \app, showing its industrial-grade applicability.
\end{itemize}

\vspace{-0.3em}
\section{Related Work}
\vspace{-0.2em}
Since Szegedy~\etal~\cite{szegedy2013intriguing} identified adversarial examples, massive adversarial attacks have been proposed to fool DNNs. White-box attacks, \eg, single-step gradient-based attack~\cite{goodfellow2014explaining}, iterative gradient-based attack~\cite{moosavi2016deepfool,kurakin2017IFGSM,madry2017towards,croce2020reliable}, and optimization-based attack~\cite{szegedy2013intriguing,carlini2017towards,athalye2018obfuscated}, often utilize the gradient and exhibit good attack performance. They have been widely adopted for evaluating the model robustness of defenses~\cite{madry2017towards,zhang2019theoretically,shafahi2019adversarial,cohen2019certified,dong2020benchmarking}, but are hard to be applied in real-world with limited information. To make adversarial attacks applicable in practice, various black-box attacks, including transfer-based attack~\cite{dong2018boosting,xie2019improving,wang2021Enhancing,wang2021admix,wu2021improving}, score-based attack~\cite{chen2017zoo,ilyas2018black,tu2019autozoom,al2019sign,du2019query,yao2019trust,zhao2020towards}, and decision-based attack~\cite{brendel2017decision,cheng2019sign,chen2020hopskipjumpattack,rahmati2020geoda,maho2021surfree}, have gained increasing interest. Among them, decision-based attack is most challenging since it can only access the prediction label. In this work, we aim to boost the query efficiency of decision-based attack by utilizing the geometric information and provide a brief overview of existing decision-based attacks.

BoundaryAttack~\cite{brendel2017decision} is the first decision-based attack that initializes a large perturbation and performs random walks on the decision boundary while keeping adversarial. Such a paradigm has been widely adopted in the subsequent decision-based attacks. OPT~\cite{cheng2018query} formulates the decision-based attack as a real-valued optimization problem with zero-order optimization. And SignOPT~\cite{cheng2019sign} further computes the sign of the directional derivative instead of the magnitude for fast convergence. HopSkipJumpAttack (HSJA)~\cite{chen2020hopskipjumpattack} boosts BoundaryAttack by estimating the gradient direction via binary information at the decision boundary. QEBA~\cite{li2020qeba} enhances HSJA for better gradient estimation using the perturbation sampled from various subspaces, including spatial, frequency, and intrinsic components. To further improve the query efficiency, qFool~\cite{liu2019geometry} assumes that the curvature of the boundary is small around adversarial examples and adopts several perturbation vectors for efficient gradient estimation. BO~\cite{shukla2021simple} uses Bayesian optimization for finding adversarial perturbations in low dimension subspace and maps it back to the original input space to obtain the final perturbation. GeoDA~\cite{rahmati2020geoda} approximates the local decision boundary by a hyperplane and searches the closest point to the benign sample on the hyperplane as the adversary. Surfree~\cite{maho2021surfree} iteratively constructs a circle on the decision boundary and adopts binary search to find the intersection of the constructed circle and decision boundary as the adversary without any gradient estimation.  

Most existing decision-based attacks restrict the adversarial example at each iteration on the decision boundary and usually adopt different gradient estimation approaches for attack. In this work, we propose Triangle Attack to minimize the adversarial perturbation in the low frequency space directly by utilizing the law of sines without gradient estimation or restricting the adversarial example on the decision boundary for efficient decision-based attack.

\section{Methodology}
In this section, we first provide the preliminaries. Then we introduce our motivation and the proposed Triangle Attack (\name). 

\subsection{Preliminaries}

Given a classifier $f$ with parameters $\theta$ and a benign sample $x \in \mathcal{X}$ with ground-truth label $y\in \mathcal{Y}$, where $\mathcal{X}$ denotes all the images and $\mathcal{Y}$ is the output space. The adversarial attack finds an adversary $x^{adv} \in \mathcal{X}$ to mislead the target model:
\begin{equation*}
\small
    f(x^{adv};\theta)\neq f(x;\theta)=y \quad\mathrm{s.t.}\quad \|x^{adv}-x\|_p <\epsilon,
\end{equation*}
where $\epsilon$ is the perturbation budget. Decision-based attacks usually first generate a large adversarial perturbation $\delta$ and then minimize the perturbation as follows:
\begin{equation}
\small
    \label{eq:goal}
    \min \|\delta\|_p \quad \mathrm{s.t.} \quad f(x+\delta;\theta)\neq f(x;\theta)=y.
\end{equation}

Existing decision-based attacks~\cite{cheng2018query,cheng2019sign,li2020qeba} often estimate the gradient to minimize the perturbation, which is time-consuming. Recently, some works adopt the geometric property to estimate the gradient or directly optimize the perturbation. Here we introduce two geometry-inspired decision-based attacks detailedly.

\textbf{GeoDA}~\cite{rahmati2020geoda} argues that the decision boundary at the vicinity of a data point $x$ can be locally approximated by a hyperplane passing through a boundary point $x_B$ close to $x$ with a normal vector $w$. Thus, Eq.~\eqref{eq:goal} can be locally linearized:
\begin{equation*}
    \min \|\delta\|_p \quad \mathrm{s.t.} \quad w^\top (x+\delta) - w^\top x_B = 0.
\end{equation*}
Here $x_B$ is a data point on the boundary, which can be found by binary search with several queries, and GeoDA randomly samples several data points for estimating $w$ to optimize the perturbation at each iteration.

\textbf{Surfree}~\cite{maho2021surfree} assumes the boundary can be locally approximated by a hyperplane around a boundary point $x+\delta$. At each iteration, it represent the adversary using polar coordinates and searches an optimal $\theta$ to update the perturbation:
% constructs a circle with the diameter from $x$ to $x+\delta$ and searches the intersection of circle and boundary as the next adversarial example. Specifically, at each iteration, it searches an optima $\theta$ to update the perturbation as follows: \TODO{polar ...}
\begin{equation*}
\small
    \delta_{t+1} = \delta_t \cos{\theta} (\bm{u}\cos{\theta}+\bm{v}\sin{\theta}),
\end{equation*}
where $\bm{u}$ is the unit vector from $x$ to $x_t^{adv}$ and $\bm{v}$ is the orthogonal vector of $\bm{u}$.

\subsection{Motivation}
\label{sec:motivation}
Different from most decision-based attacks with gradient estimation~\cite{cheng2018query,cheng2019sign,li2020qeba,rahmati2020geoda} or random walk on the decision boundary~\cite{brendel2017decision,maho2021surfree}, we aim to optimize the perturbation using the geometric property without any queries for gradient estimation. After generating a large adversarial perturbation, the decision-based attacks move the adversarial example close to the benign sample, \ie, decrease the adversarial perturbation $\delta_t$, while keeping the adversarial property at each iteration. In this work, as shown in Fig.~\ref{fig:TA}, we find that at the $t$-th iteration, the benign sample $x$, current adversarial example $x_t^{adv}$ and next adversarial example $x_{t+1}^{adv}$ can naturally construct a triangle in a subspace for any iterative attacks. Thus, searching for the next adversarial example $x_{t+1}^{adv}$ with smaller perturbation is equivalent to searching for a triangle based on $x$ and $x_t^{adv}$, in which the third data point $x'$ is adversarial and satisfies $\|x'-x\|_p < \|x_t^{adv}-x\|_p$. This inspires us to utilize the relationship between the angle and side length in the triangle to search an appropriate triangle to minimize the perturbation at each iteration. As shown in Sec.~\ref{sec:exp:ablation}, however, directly applying such a geometric property on the input image leads to very poor performance. Thanks to the generality of such a geometric property, we optimize the perturbation in the low frequency space generated by DCT~\cite{ahmed1974discrete} for effective dimensionality reduction, which exhibits great attack efficiency as shown in Sec.~\ref{sec:exp:ablation}.

Moreover, since Brendel~\etal~\cite{brendel2017decision} proposed BoundaryAttack, most decision-based attacks~\cite{cheng2018query,cheng2019sign,chen2020hopskipjumpattack,rahmati2020geoda,maho2021surfree} follow the setting in which the adversarial example at each iteration should be on the decision boundary. We argue that such a restriction is not necessary in decision-based attacks but introduces too many queries on the target model to approach the boundary. Thus, we do not adopt this constraint in this work and validate this argument in Sec.~\ref{sec:exp:ablation}.

\subsection{Triangle Attack}
\label{sec:method}
In this work, we have the following assumption that the adversarial examples exist for any deep neural classifier $f$: 
\vspace{-0.5em}
\begin{Assumption}
    Given a benign sample $x$ and a perturbation budget $\epsilon$, there exists an adversarial perturbation $\|\delta\|_p \leq \epsilon$ towards the decision boundary which can mislead the target classifier $f$.
    \vspace{-0.5em}
\end{Assumption}

This is a general assumption that we can find the adversarial example $x^{adv}$ for the input sample $x$, which has been validated by numerous works~\cite{goodfellow2014explaining,carlini2017towards,athalye2018obfuscated,brendel2017decision,wang2021boosting}. If this assumption does not hold, the target model is ideally robust so that we cannot find any adversarial example within the perturbation budget, which is beyond our discussion. Thus, we follow the framework of existing decision-based attacks by first randomly crafting a large adversarial perturbation and then minimizing the perturbation. To align with previous works, we generate a random perturbation close to the decision boundary with binary search~\cite{li2020qeba,rahmati2020geoda,maho2021surfree} and mainly focus on the perturbation optimization.

In two arbitrary consecutive iterations of the perturbation optimization process for any adversarial attacks, namely $t$-th and $(t+1)$-th iterations without loss of generalization, the input sample $x$, current adversarial example $x_t^{adv}$ and the next adversarial example $x_{t+1}^{adv}$ can naturally construct a triangle in a subspace of the input space $\mathcal{X}$. Thus, as shown in Fig.~\ref{fig:TA}, decreasing the perturbation to generate $x_{t+1}^{adv}$ is equivalent to searching for an appropriate triangle in which the three vertices are $x$, $x_t^{adv}$ and $x_{t+1}^{adv}$, respectively.

\vspace{-0.5em}
\begin{theorem}[The law of sines] 
Suppose $a$, $b$ and $c$ are the side lengths of a triangle, and $\alpha$, $\beta$ and $\gamma$ are the opposite angles, we have $\frac{a}{\sin{\alpha}}=\frac{b}{\sin{\beta}}=\frac{c}{\sin{\gamma}}$.
\label{thm:sines}
\vspace{-0.5em}
\end{theorem}

From Theorem~\ref{thm:sines}, we can obtain the relationship between the side length and opposite angle for the triangle in Fig.~\ref{fig:TA}:
\begin{equation}
\small
    \frac{\delta_t}{\sin{\alpha_t}} = \frac{\delta_{t+1}}{\sin{(\pi - (\alpha_t + \beta_t))}}.
\end{equation}
To greedily decrease the perturbation $\delta_t$, the $t$-th triangle should satisfy that $\frac{\delta_{t+1}}{\delta_t}=\frac{\sin{(\pi - (\alpha_t + \beta_t))}}{\sin{\alpha_t}}<1$, \ie, $\pi - (\alpha_t + \beta_t) < \alpha_t$. Thus, decreasing the perturbation at the $t$-th iteration can be achieved by finding a triangle constructed by the input sample $x$, current adversarial example $x_t^{adv}$ and the angles $\beta_t$ and $\alpha_t$, which satisfy $\beta_t + 2\alpha_t > \pi$ and the third vertex should be adversarial. We denote such a triangle as candidate triangle and $\mathcal{T}(x, x_{t}^{adv}, \alpha_t, \beta_t, \mathcal{S}_t)$ as the third vertex, where $\mathcal{S}_t$ is a sampled subspace. Based on this observation, we propose a novel decision-based attack called Triangle Attack (\name) that searches the candidate triangle at each iteration and adjusts the angle $\alpha_t$ accordingly. 

\begin{wrapfigure}{R}{0.435\textwidth}
  \centering
  \vspace{-2.2em}
    \includegraphics[width=\linewidth]{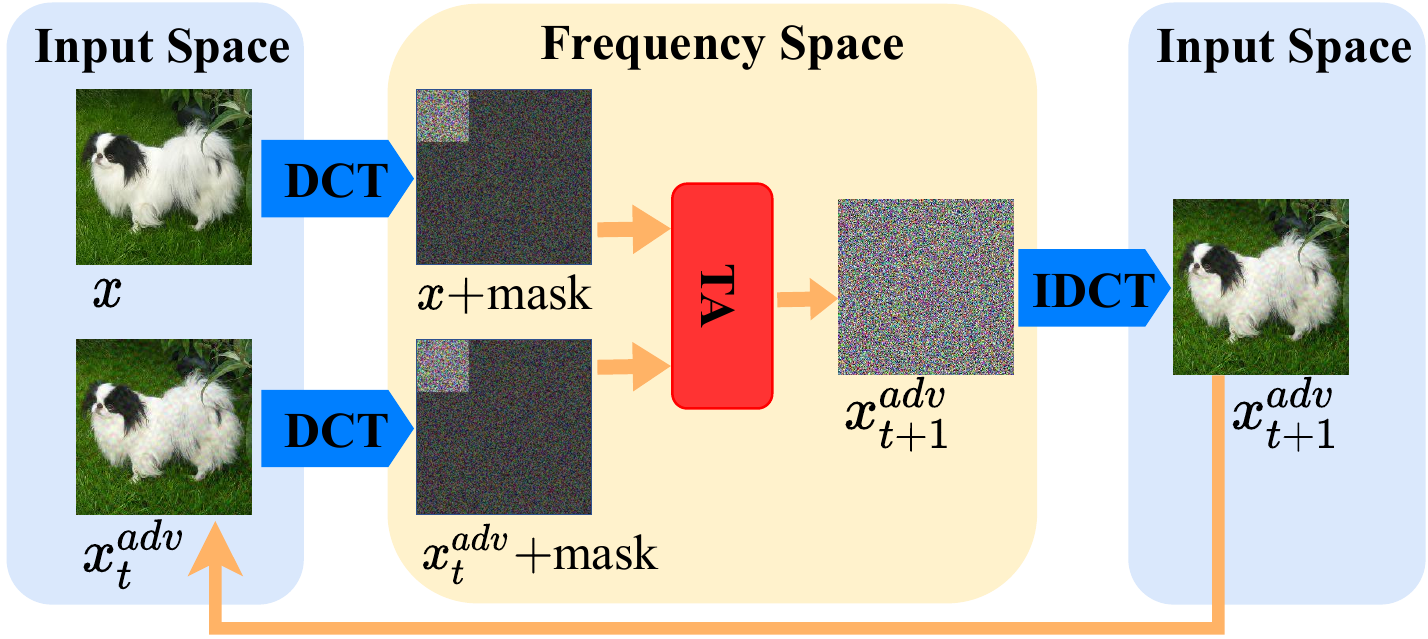}
   \caption{Illustration of the entire procedure of \name attack at the $t$-th iteration. We construct the triangle in the frequency space to efficiently craft adversarial examples. Note that here we adopt DCT for illustration but we do not need it for $x$ at each iteration. We still adopt $x$ and $x_t^{adv}$ in the frequency space without ambiguity due to the one-to-one mapping of DCT}
  \label{fig:freq}
  \vspace{-2.2em}
\end{wrapfigure}

\textbf{Sampling the 2-D subspace $\mathcal{S}$ of frequency space.} The input image often lies in a high-dimensional space, such as $224\times224\times3$ for ImageNet~\cite{krizhevsky2012imagenet}, which is too large for the attack to explore the neighborhood for minimizing the adversarial perturbation efficiently. Previous works~\cite{guo2018low,li2020qeba,maho2021surfree} have shown that utilizing the information in various subspaces can improve the efficiency of decision-based attacks. For instance, QEBA~\cite{li2020qeba} samples the random noise for gradient estimation in the spatial transformed space or low frequency space but minimizes the perturbation in the input space with the estimated gradient. Surfree~\cite{maho2021surfree} optimizes the perturbation in the subspace of the input space determined by a unit vector randomly sampled in the low frequency space. In general, the low frequency space contains the most critical information for image. With the poor performance of \name in the input space as shown in Sec.~\ref{sec:exp:ablation} and the generality of the geometric property as shown in Fig.~\ref{fig:freq}, we directly optimize the perturbation in the frequency space at each iteration for effective dimensionality reduction. And we randomly sample a $d$-dimensional line across the benign sample in the low frequency space (top 10\%). The sampled line, directional line from benign sample $x$ and current adversary $x_t^{adv}$ can determine a unique 2-D subspace $\mathcal{S}$ of the frequency space, in which we can construct the candidate triangle to minimize the perturbation. The final adversary can be converted into the input space by Inverse DCT (IDCT).
% Note that different from existing attacks (\eg QEBA, Surfree) that optimizes the perturbation in the input space with the help of low frequency components, TA directly optimizes the perturbation in the frequency space, which significantly boosts the query efficiency as shown in Sec.~\ref{sec:exp:ablation}.

% \input{figs/angle_effect}

\begin{figure}[t]
    \begin{minipage}[c]{0.5\linewidth}
        \centering
        \includegraphics[width=0.93\linewidth]{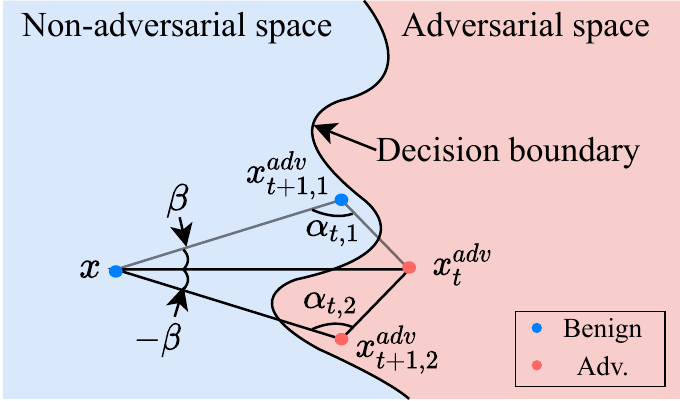}
        \caption{Illustration of a symmetric candidate triangle ($x$, $x_t^{adv}$ and $x_{t+1,2}^{adv}$). When the angle $\beta$ cannot result in adversarial example ($x_{t+1,1}^{adv}$), we would further construct the symmetric triangle based on the line $\langle x, x_t^{adv} \rangle$ to check the data point $x_{t+1,2}^{adv}$}
        \label{fig:symmetry}
    \end{minipage}%
    \hspace{1em}
    \begin{minipage}[c]{0.48\linewidth}
        \centering
        \includegraphics[width=0.97\linewidth]{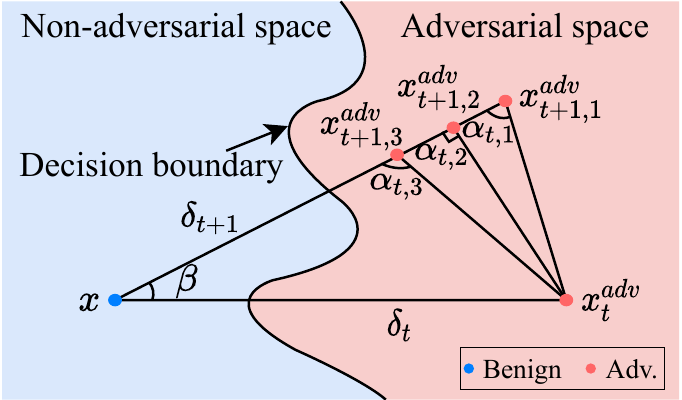}
        \caption{The effect of the magnitude of $\alpha$ for the candidate triangle used in \name. For the same sampled angle $\beta$, the larger angle $\alpha$ leads to smaller perturbation but is also more likely to cross over the decision boundary}
        \label{fig:angle_effect}
    \end{minipage}%
\vspace{-1.3em}
\end{figure}

\textbf{Searching the candidate triangle.} Given a subspace $\mathcal{S}_t$, the candidate triangle only depends on the angle $\beta$ since $\alpha$ is updated during the optimization. As shown in Fig.~\ref{fig:symmetry}, if we search an angle $\beta$ without leading to an adversarial example ($x_{t+1,1}^{adv}$), we can further construct a symmetric triangle with the same angle in the opposite direction to check the data point $x_{t+1,2}^{adv}$, which has the same magnitude of perturbation as $x_{t+1,1}^{adv}$ but different directions. We denote the angle as $-\beta$ for the symmetric triangle without ambiguity. Note that with the same angle $\alpha$, a larger angle $\beta$ would make the third vertex closer to the input sample $x$, \ie, smaller perturbation. After determining the subspace $\mathcal{S}_t$, we first check the angle $\beta_{t,0}=\max(\pi-2\alpha, \underline{\beta})$, where $\underline{\beta}=\pi/16$ is a pre-defined small angle. If neither $\mathcal{T}(x,x_t^{adv},\alpha_t, \beta_{t,0},\mathcal{S}_t)$ nor $\mathcal{T}(x,x_t^{adv},\alpha_t, -\beta_{t,0},\mathcal{S}_t)$ is adversarial, 
we give up this subspace because it cannot bring any benefit. Otherwise, we adopt binary search to find an optimal angle $\beta^\ast \in [\max(\pi-2\alpha, \underline{\beta}), \min(\pi-\alpha, \pi/2)]$ which is as large as possible to minimize the perturbation. Here we restrict the upper bound of $\beta$ because $\mathcal{T}(x,x_t^{adv},\alpha_t, \beta,\mathcal{S}_t)$ would be at the opposite direction \wrt $x$ for $\beta > \pi/2$ and $\pi-\alpha$ guarantees a valid triangle.

\textbf{Adjusting the angle $\alpha$.} Intuitively, the angle $\alpha$ balances the magnitude of perturbation and the difficulty to find an adversarial example:
\vspace{-0.5em}
\begin{proposition}
    With the same angle $\beta$, a smaller angle $\alpha$ makes it easier to find an adversarial example while a larger angle $\alpha$ leads to smaller perturbation.
\end{proposition}
\vspace{-0.5em}
\noindent Intuitively, as shown in Fig.~\ref{fig:angle_effect}, the smaller angle $\alpha$ results in larger perturbation but is more likely to cross over the decision boundary, making it easier to search an adversarial example, and vice versa. It is hard to consistently find an optimal $\alpha$ for each iteration, letting alone various input images and target models. Thus, we adaptively adjust the angle $\alpha$ based on the crafted adversarial example:
\begin{equation}
\small
    \alpha_{t,i+1} = \left\{ \begin{array}{cc}
        \min(\alpha_{t,i}+\gamma, \pi/2+\tau) & \mathrm{if} \  f(x_{t,i+1}^{adv};\theta)\neq y\\
        \max(\alpha_{t,i}-\lambda\gamma, \pi/2-\tau) & \mathrm{Otherwise}
    \end{array} \right.
    \label{eq:alpha}
\end{equation}
where $x_{t,i+1}^{adv}=\mathcal{T}(x,x_t^{adv},\alpha_{t,i},\beta_{t,i},\mathcal{S}_t)$ is the adversarial example generated by $\alpha_{t,i}$, $\gamma$ is the change rate, $\lambda$ is a constant, and $\tau$ restricts the upper and lower bounds of $\alpha$. We adopt $\lambda < 1$ to prevent decreasing the angle too fast considering much more failures than successes during the perturbation optimization. Note that the larger angle $\alpha$ makes it harder to find an adversarial example. However, a too small angle $\alpha$ results in a much lower bound for $\beta$, which also makes $\mathcal{T}(x,x_t^{adv}, \alpha_t, \beta_t, \mathcal{S}_t)$ far away from the current adversarial example $x_t^{adv}$, decreasing the probability to find an adversarial example. Thus, we add the bounds for $\alpha$ to restrict it in an appropriate range.

\begin{algorithm}[t] 
    \small
    \newcommand\mycommfont[1]{\footnotesize\ttfamily{#1}}
    \SetCommentSty{mycommfont}
    \caption{Triangle Attack}
	\label{alg:alg_ta}
	\KwIn{Target classifier $\bm{f}$ with parameters $\bm{\theta}$; Benign sample $\bm{x}$ with ground-truth label $\bm{y}$; Maximum number of queries $Q$; Maximum number of iteration $N$ for each sampled subspace; Dimension of the directional line $d$; Lower bound $\underline{\beta}$ for angle $\beta$.}
	\KwOut{An adversarial example $x^{adv}$.}  
	Initialize a large adversarial perturbation $\delta_0$;\\
	$x_0^{adv}=x+\delta_0$, $q=0$, $t=0$, $\alpha_0 = \pi/2$;\\
    \While{$q<Q$}{\label{alg:line:restart}
        Sampling 2-D subspace $\mathcal{S}_t$ in the low frequency space; \\
        $\beta_{t,0}=\max(\pi-2\alpha,\underline{\beta})$;\\
        \If{$f(\mathcal{T}(x, x_t^{adv},\alpha_{t,0}, \beta_{t,0}, \mathcal{S}_t);\theta)=f(x;\theta)$}{
                $q= q+1$, update $\alpha_{t,0}$ based on Eq.~\eqref{eq:alpha};\\
                 \If{$f(\mathcal{T}(x, x_t^{adv}, \alpha_{t,0},-\beta_{t,0}, \mathcal{S}_t);\theta)=f(x;\theta)$}{
                    $q=q+1$, update $\alpha_{t,0}$ based on Eq.~\eqref{eq:alpha}; \\
                    Go to line~\ref{alg:line:restart}; \algorithmiccomment{give up this subspace}
                }
            }
        $\overline{\beta}_{t,0} = \min(\pi/2, \pi-\alpha)$;\\
        \For(\Comment{binary search for angle $\beta$}){$i=0 \to N$}{
                $\beta_{t, i+1} = (\overline{\beta}_{t,i}+\beta_{t,i})/2$;\\
                \If{$f(\mathcal{T}(x, x_t^{adv}, \alpha_{t,i}, \beta_{t, i+1}, \mathcal{S}_t);\theta) = f(x;\theta)$}{
                $q = q+1$, update $\alpha_{t,i}$ based on Eq.~\eqref{eq:alpha};\\
                    \If{$f(\mathcal{T}(x, x_t^{adv}, \alpha_{t,i}, -\beta_{t, i+1}, \mathcal{S}_t);\theta) = f(x;\theta)$}{
                        $\overline{\beta}_{t,i+1} = \beta_{t,i+1}, \beta_{t,i+1}=\beta_{t,i}$;\\
                    }
                }
                $q=q+1$, update $\alpha_{t,i+1}$ based on Eq.~\eqref{eq:alpha};\\
            }
    $x_{t+1}^{adv}=\mathcal{T}(x, x_t^{adv}, \alpha_{t,i+1}, \beta_{t, i+1}, \mathcal{S}_t)$, $t=t+1$; \\
    }
    \Return $x_t^{adv}$.\\
\end{algorithm}

\name iteratively searches the candidate triangle in the subspace $\mathcal{S}_t$ sampled from the low frequency space to find the adversarial example and update the angle $\alpha$ accordingly. The entire algorithm of \name is summarized in Algorithm~\ref{alg:alg_ta}.

\section{Experiments}

In this section, we conduct extensive evaluations on the standard ImageNet dataset using five models and \app to evaluate the effectiveness of \name. Code is available at \url{https://github.com/xiaosen-wang/TA}.
% To validate the effectiveness of the proposed \name, we conduct extensive evaluations on the standard ImageNet dataset using five models and \app. In this section, we first specify the experimental setup, and then we compare \name with various decision-based attacks on offline and online models. Experimental results demonstrate that \name can achieve much higher attack success rate within 1,000 queries and needs much fewer number of queries to achieve the same attack success rate than existing decision-based attacks. Finally, we further provide ablation studies on the hyper-parameters and further discussions.

% To validate the effectiveness of \name, we conduct extensive evaluations using standard ImageNet dataset~\cite{russakovsky2015imagenet} on various popular models~\cite{simonyan2014very,szegedy2016inceptionv3,he2016deep,huang2017densely}. Experimental results demonstrate that \name can exhibit much higher attack success rate within 1,000 queries and needs much lower number of queries when achieving the same attack success rate under the same perturbation budget on various models than existing decision-based attacks~\cite{cheng2018query,cheng2019sign,chen2020hopskipjumpattack,li2020qeba,rahmati2020geoda,maho2021surfree}. With the high query efficiency, we further validate the practical applicability on \app, in which \name can successfully generate more adversarial examples with imperceptible perturbation as shown in Sec.~\ref{sec:realworld}.

\vspace{-0.2em}
\subsection{Experimental Setup}

\textbf{Dataset.} To validate the effectiveness of the proposed \name, following the setting of Surfree~\cite{russakovsky2015imagenet}, we randomly sample 200 correctly classified images from the ILSVRC 2012 validation set for evaluation on the corresponding models. 

\textbf{Models.} We consider five widely adopted models, \ie, VGG-16~\cite{simonyan2014very}, Inception-v3~\cite{szegedy2016rethinking}, ResNet-18~\cite{he2016deep}, ResNet-101~\cite{he2016deep} and DenseNet-121~\cite{huang2017densely}. To validate the applicability in the real world, we evaluate \name on \app\footnote{https://cloud.tencent.com/}.

\textbf{Baselines.} We take various decision-based attacks as our baselines, including four gradient estimation based attacks, \ie, OPT~\cite{cheng2018query}, SignOPT~\cite{cheng2019sign}, HSJA~\cite{chen2020hopskipjumpattack}, QEBA~\cite{li2020qeba}, one optimization based attack, \ie, BO~\cite{shukla2021simple}, and two geometry-inspired attacks, \ie, GeoDA~\cite{rahmati2020geoda}, Surfree~\cite{maho2021surfree}.

\textbf{Evaluation metrics.} Following the standard setting in QEBA~\cite{li2020qeba}, we adopt the root mean squared error ($RMSE$) between benign sample $x$ and adversarial example $x^{adv}$ to measure the magnitude of perturbation:
\begin{equation}
\small
d(x, x^{adv})=\sqrt{\frac{1}{w\cdot h\cdot c}\sum_{i=1}^w \sum_{j=1}^h \sum_{k=1}^c (x[i,j,k]-x^{adv}[i,j,k])^2},
\end{equation}
where $w,h,c$ are the width, height and number of channel of the input image, respectively. We also adopt the attack success rate, the percentage of adversarial examples which reach a certain distance threshold. 

\textbf{Hyper-parameters.} For fair comparison, all the attacks adopt the same adversarial perturbation initialization approach as in \cite{maho2021surfree} and the hyper-parameters for baselines are exactly same as in the original papers. For our \name, we adopt the maximum number of iteration in each subspace $N=2$, the dimension of directional line $d=3$ and $\gamma=0.01$, $\lambda=0.05$ and $\tau=0.1$ for updating angle $\alpha$.

% \begin{table*}
% % \small
%   \centering
%   \begin{tabular}{lccccccccccccccc}
%     \toprule
%     \multirow{2}{*}{Method} & \multicolumn{3}{c}{Inception-v3} & \multicolumn{3}{c}{VGG-16} & \multicolumn{3}{c}{ResNet-18}  & \multicolumn{3}{c}{ResNet-101 }  & \multicolumn{3}{c}{DenseNet-121} \\
%     \cmidrule(lr){2-4} \cmidrule(lr){5-7} \cmidrule(lr){8-10} \cmidrule(lr){11-13} \cmidrule(lr){14-16} 
%     & ASR & $\|\delta\|_2$ & AQ & ASR & $\|\delta\|_2$ & AQ & ASR & $\|\delta\|_2$ & AQ  & ASR & $\|\delta\|_2$ & AQ & ASR & $\|\delta\|_2$ & AQ\\
%     \midrule
%     OPT & 13.5  & 66.9 & 1094 & 35.5 & 25.5 & 1132 & 29.5 & 27.9 & 1122 & & & & & & \\
%     SignOPT & 18.5 & 46.3 & 1115 & 58.5 & 16.2  & 1147 & 51.5 & 18.9  & 1148 & & & & & & \\
%     HSJA & 46.5 & 19.2 & 1017 & 62.5 & 14.4 & 1014 & 75.5 & 10.5 & 1014 & & & & & & \\
%     QEBA & 30.5 & 44.4 & 1056 & 63.5 & 14.3 &  1054& 59.5 & 15.8 & 1054 & & & & & & \\
%     Surfree & 49.0 & 46.3 & 1004  & 80.5 &8.7 &1003 &77.5 & 9.7& 1004 & & & & & & \\
%     \name & \textbf{49.5} & \textbf{18.8} & \textbf{1001} & \textbf{93.0} & \textbf{6.21} & \textbf{1001} & \textbf{82.0} & \textbf{7.6} & \textbf{1001} & & &  & & &  \\
%     \bottomrule
%   \end{tabular}
%   \caption{Results.   Ours is better.}
%   \label{tab:example}
% \end{table*}

\begin{table*}[t]
% \small
  \centering
    \caption{Attack success rate (\%) on five models under different RMSE thresholds. The maximum number of queries is set to 1,000. We highlight the highest attack success rate in \textbf{bold}}
    \label{tab:main_result}
  \resizebox{\textwidth}{!}{
  \begin{tabular}{l>{\rowmac}c>{\rowmac}c>{\rowmac}c>{\rowmac}c>{\rowmac}c>{\rowmac}c>{\rowmac}c>{\rowmac}c>{\rowmac}c>{\rowmac}c>{\rowmac}c>{\rowmac}c>{\rowmac}c>{\rowmac}c>{\rowmac}c}
    \toprule
    Model & \multicolumn{3}{c}{VGG-16} & \multicolumn{3}{c}{Inception-v3} & \multicolumn{3}{c}{ResNet-18}  & \multicolumn{3}{c}{ResNet-101 }  & \multicolumn{3}{c}{DenseNet-121} \\
    \cmidrule(lr){2-4} \cmidrule(lr){5-7} \cmidrule(lr){8-10} \cmidrule(lr){11-13} \cmidrule(lr){14-16} 
   RMSE & 0.1 & 0.05 & 0.01 &  0.1 & 0.05 & 0.01 &  0.1 & 0.05 & 0.01 &  0.1 & 0.05 & 0.01 &  0.1 & 0.05 & 0.01\\
    \midrule
    OPT  &~~76.0 & 38.5 & ~~5.5& 34.0 & 17.0 & ~~4.0 & ~~67.0 & 36.0 & ~~6.0  & 51.5 & 21.0 & ~~5.0 & ~~51.5 & 29.0 & ~~5.5 \\
    SignOPT  & ~~94.0 & 57.5 & 12.5& 50.5 & 27.0 & ~~8.0 & ~~84.5 & 49.5 & 13.0  & 69.0 & 33.0 & ~~8.0 & ~~69.5 & 44.0 & 10.0 \\
    HSJA  & ~~92.5 & 58.5 & 13.0& 32.5 & 14.0 & ~~4.0 & ~~83.0 & 51.0 & 12.5  & 71.5 & 37.5 & 12.0 & ~~70.5 & 43.5 & 10.5 \\
    QEBA &  ~~98.5 & 86.0 & 29.0& 78.5 & 54.5 & 17.0 & ~~98.0 & 81.5 & 34.5  & 94.0 & 59.0 & 20.5 & ~~91.0 & 66.0 & 24.0 \\
    BO &  ~~96.0 & 72.5 & 17.0 & 75.5 & 43.0 & 10.0 & ~~94.5 & 74.0 & 16.0 & 89.5 & 63.0 & 16.5 & ~~93.0 & 64.5 & 16.5 \\
    GeoDA & ~~99.0 & 94.0 & 35.0& 89.0 & 61.5 & 23.5 & ~~99.5 & 90.0 & 30.5  & 98.0 & 81.5 & 22.0 & \bf{100.0} & 84.5 & 27.5 \\
    Surfree  & ~~99.5 & 92.5 & 39.5& 87.5 & 67.5 & 24.5 & ~~98.5 & 87.0 & 36.0  & 95.5 & 76.5 & 27.0 & ~~97.0 & 78.0 & 29.0 \\
      \name(\bf{Ours})  &\bf{100.0} & \bf{95.0} & \bf{44.5}& \bf{96.5} & \bf{81.5} & \bf{30.0} & \bf{100.0} & \bf{94.0} & \bf{51.5}  & \bf{99.0} & \bf{88.5} & \bf{40.0} & ~~99.5 & \bf{92.5} & \bf{43.5}   \clearrow\\
    \bottomrule
  \end{tabular}}
\end{table*}

\vspace{-0.2em}
\subsection{Evaluation on Standard Models}
To evaluate the effectiveness of \name, we first compare the attack performance on five popular models with different decision-based attacks and report the attack success rate under various $RMSE$ thresholds, namely 0.1, 0.05 and 0.001. 

\begin{figure*}[t]
    \begin{subfigure}{.315\linewidth}
    \centering
    \includegraphics[width=\linewidth]{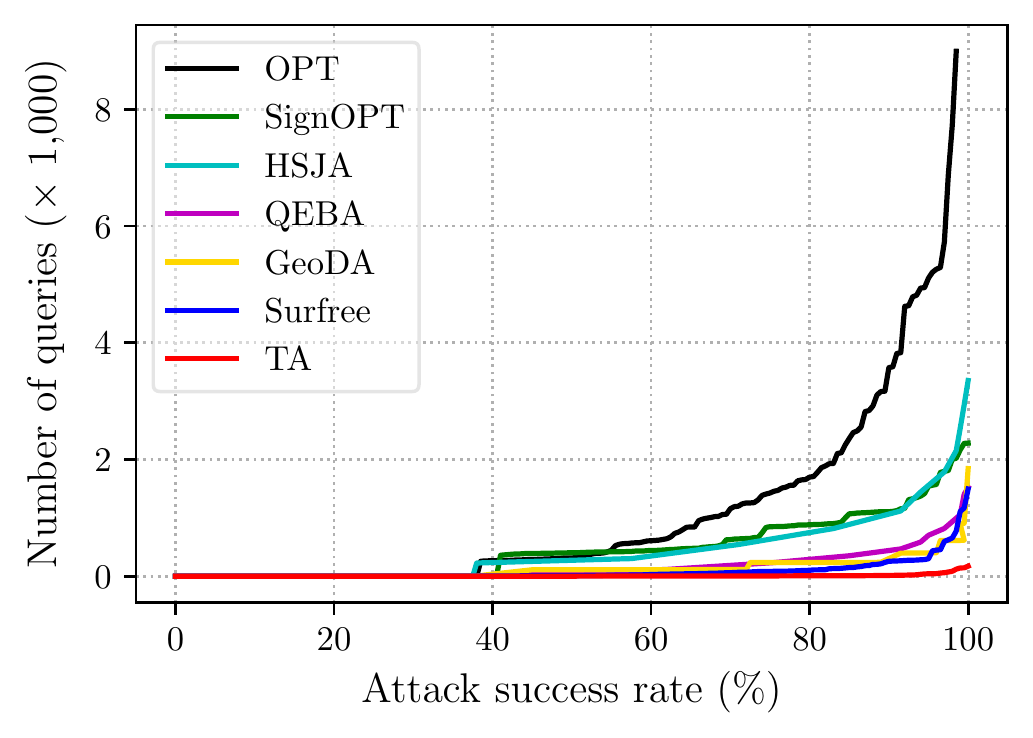}
    \caption{RMSE=0.1}
    \label{fig:query_succ:0.1}
    \end{subfigure}%
    \hfill
    \begin{subfigure}{.315\linewidth}
    \centering
    \includegraphics[width=\linewidth]{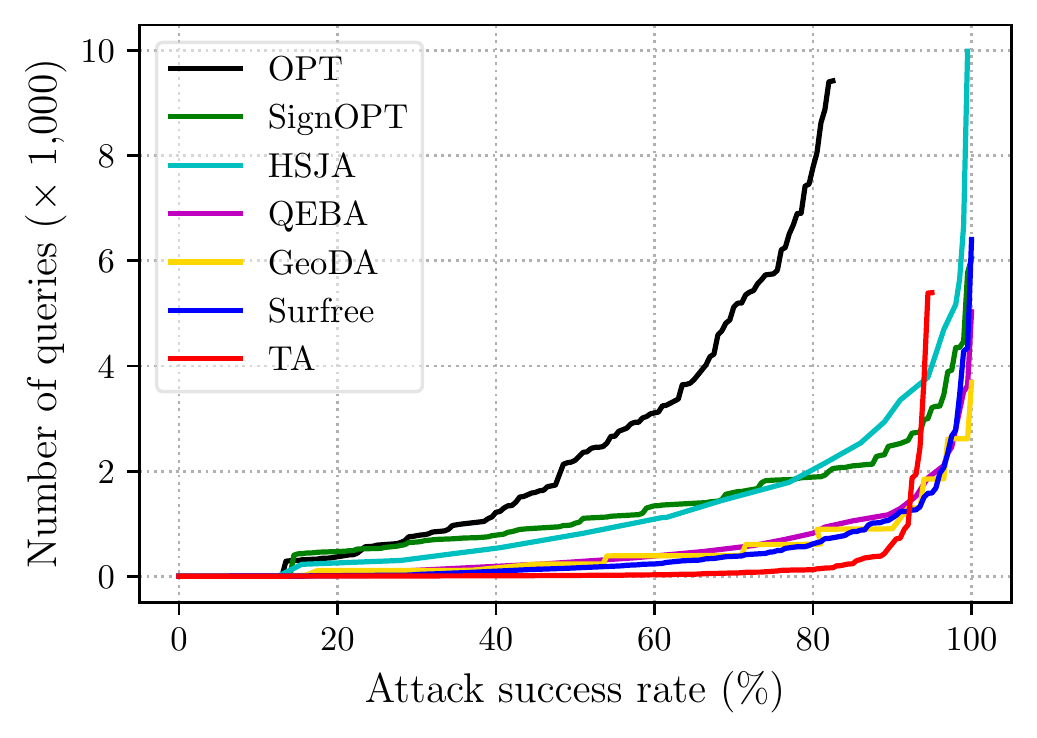}
    \caption{RMSE=0.05}
    \label{fig:query_succ:0.05}
    \end{subfigure}%
    \hfill
    \begin{subfigure}{.315\linewidth}
    \centering
    \includegraphics[width=\linewidth]{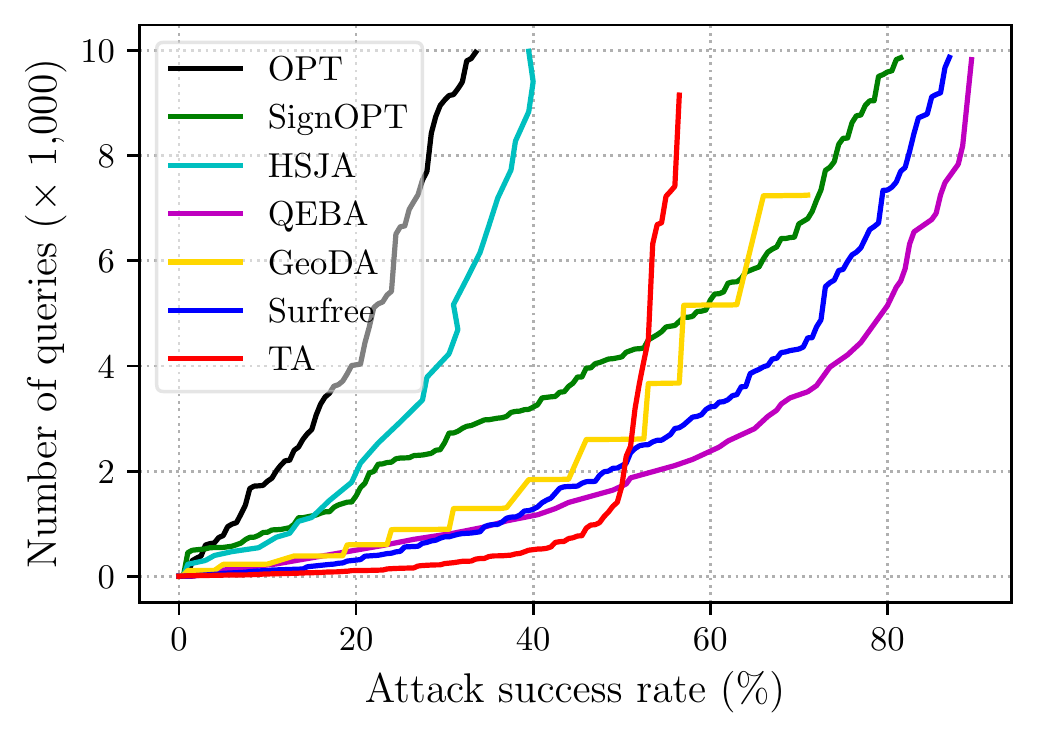}
    \caption{RMSE=0.01}
    \label{fig:query_succ:0.01}
    \end{subfigure}
    \vspace{-0.7em}
    \caption{Number of queries to achieve the given attack success rate on ResNet-18 for the attack baselines and the proposed \name under various perturbation budgets. The maximum number of queries is 10,000}
    \label{fig:query_succ}
    \vspace{-0.5em}
\end{figure*}

We first evaluate the attack within 1,000 queries, which is widely adopted in recent works~\cite{chen2020hopskipjumpattack,rahmati2020geoda,maho2021surfree}.
The attack success rate is summarized in Table~\ref{tab:main_result}, which means the attack would fail to generate adversarial example for the input image if it takes 1,000 queries without reaching the given threshold. We can observe that \name consistently achieves much better attack success rate than existing decision-based attacks under various perturbation budgets on five models with different architectures. For instance, \name outperforms the runner-up attack with a clear margin of 1.0\%, 7.5\% and 13.0\% under the $RMSE$ threshold of 0.1, 0.05, 0.01 on ResNet-101, which is widely adopted for evaluating the decision-based attacks. In particular, the proposed \name significantly outperforms the two geometry-inspired attacks, \ie, GeoDA~\cite{rahmati2020geoda} and Surfree~\cite{maho2021surfree}, which exhibit the best attack performance among the baselines. This convincingly validates the high effectiveness of the proposed \name. Besides, among the five models, Inception-v3~\cite{szegedy2016inceptionv3}, which is rarely investigated in decision-based attacks, exhibits better robustness than other models under various perturbation budgets against both baselines and \name. Thus, it is necessary to thoroughly evaluate the decision-based attacks on various architectures instead of only ResNet models. 

To further verify the high efficiency of \name, we investigate the number of queries to achieve various attack success rates under the $RMSE$ threshold of 0.1, 0.05 and 0.01, respectively. The maximum number of queries is set to 10,000 and the results on ResNet-18 are summarized in Fig.~\ref{fig:query_succ}. As shown in Fig.~\ref{fig:query_succ:0.1} and \ref{fig:query_succ:0.05}, \name needs much less number of queries to achieve various attack success rates with $RMSE$ threshold of 0.1 and 0.05, showing the high query efficiency of our method. For the smaller threshold of 0.01, as shown in Fig.~\ref{fig:query_succ:0.01}, our \name still needs less number of queries when achieving the attack success rate smaller than 50\% but fails to achieve the attack success rate higher than 60\%. Note that as shown in Fig.~\ref{fig:tencent_res} and Table~\ref{tab:main_result}, $RMSE$ threshold of 0.01 is very rigorous so that the perturbation is imperceptible but is also hard to generate the adversarial examples for decision-based attacks. Since we mainly focus on the query efficiency of attack only based on the geometric information, the attack performance under the $RMSE$ threshold of 0.01 is acceptable because it is impractical for such high number of queries when attacking real-world applications. 

Besides, since \name aims to improve the query efficiency by utilizing the triangle geometry, the global optima might be worse than existing gradient estimation based attacks when more queries are allowed. Other geometry-inspired methods also perform poorer than QEBA~\cite{li2020qeba} in this case without gradient estimation. However, it is not the goal of \name and can be easily solved using gradient estimation. With the high efficiency of \name, we can achieve higher attack performance with lower number of queries by taking the \name as warm-up for the precise gradient estimation attacks, such as QEBA~\cite{li2020qeba}, if the high number of queries is acceptable. We integrate the gradient estimation used in QEBA~\cite{li2020qeba} into \name after 2,000 queries, dubbed \textbf{TAG}. For the perturbation budget of 0.01, TAG achieves the attack success rate of 95\% using 7,000 queries, which is better than the best baseline with the attack success rate of 92\% using 9,000 queries.

\subsection{Evaluation on Real-world Applications}
\label{sec:realworld}
With the superior performance and unprecedented progress of DNNs, numerous companies have deployed DNNs for a variety of tasks and also provide commercial APIs (Application Programming Interfaces) for different tasks. Developers can pay for these services to integrate the APIs into their applications. However, the vulnerability of DNNs to adversarial examples, especially the prosperity of decision-based attack which does not need any information of target models, poses severe threats to these real-world applications. With the high efficiency of \name, we also validate its practical attack applicability using \app. Due to the high cost of commercial APIs, we randomly sample 20 images from ImageNet validation set and the maximum number of queries is 1,000.

\begin{table*}[t]
    \centering
    \vspace{-1.5em}
     \caption{The number of adversarial examples successfully generated by various attack baselines and the proposed \name on \app within 200/500/1,000 queries. The results are evaluated on 20 randomly sampled images from the correctly classified images in ImageNet due to the high cost of online APIs}
    \label{tab:tencent_api}
    \begin{tabular}{lccccccc}
        \toprule
        RMSE & OPT & SignOPT & HSJA & QEBA & GeoDA & Surfree & \name (\textbf{Ours}) \\
        \midrule
        0.1 & 4/6/6 & 8/8/9 & 7/8/8 & 12/12/12 & 15/15/15 & 13/13/13 & \bf{17/17/17}\\
        0.05 & 1/3/3 & 4/4/7 & 6/6/8 & 11/11/12 & 13/14/14 & 12/12/13 & \bf{15/17/17}\\
        0.01 & 1/1/2 & 1/1/3 & 2/5/6 & 3/8/9 & 3/7/12 & 5/8/10 & \bf{8/12/13}\\
        \bottomrule
    \end{tabular}
    \vspace{-0.5em}
\end{table*}

\begin{figure}[t]
    \centering
    \hspace{15pt}
    \begin{minipage}[c]{0.133\textwidth} 
        \caption*{Benign Sample}
        \vspace{-4pt}
        \begin{subfigure}{\textwidth}
          \centering 
          \makebox[0pt][r]{\makebox[20pt]{\raisebox{20pt}{\rotatebox[origin=c]{90}{\small Bird}}}}%
          \hspace{-8pt}
          \includegraphics[width=\linewidth]{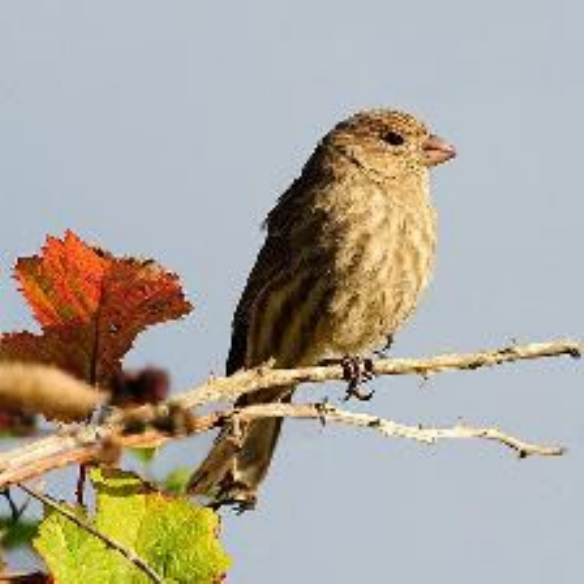}
          \vspace{13pt}
        \end{subfigure}\\
        \begin{subfigure}{\textwidth} 
          \centering 
          \makebox[0pt][r]{\makebox[20pt]{\raisebox{20pt}{\rotatebox[origin=c]{90}{\small Sea}}}}%
          \hspace{-8pt}
          \includegraphics[width=\linewidth]{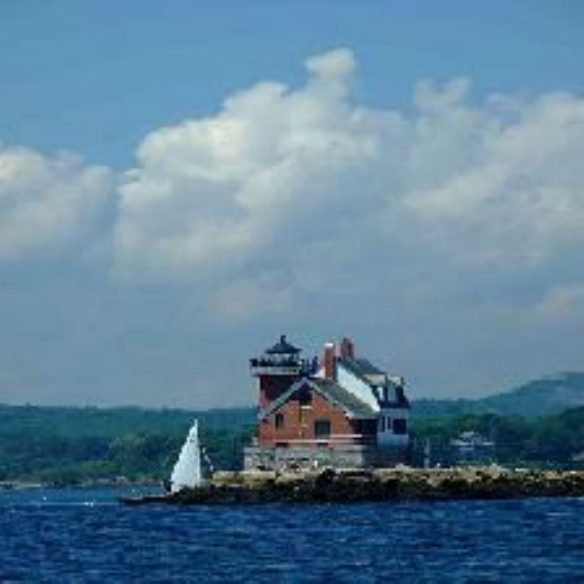}
          \vspace{13pt}
        \end{subfigure}\\
        \begin{subfigure}{\textwidth} 
          \centering 
          \makebox[0pt][r]{\makebox[20pt]{\raisebox{20pt}{\rotatebox[origin=c]{90}{\small Koala}}}}%
          \hspace{-8pt}
          \includegraphics[width=\linewidth]{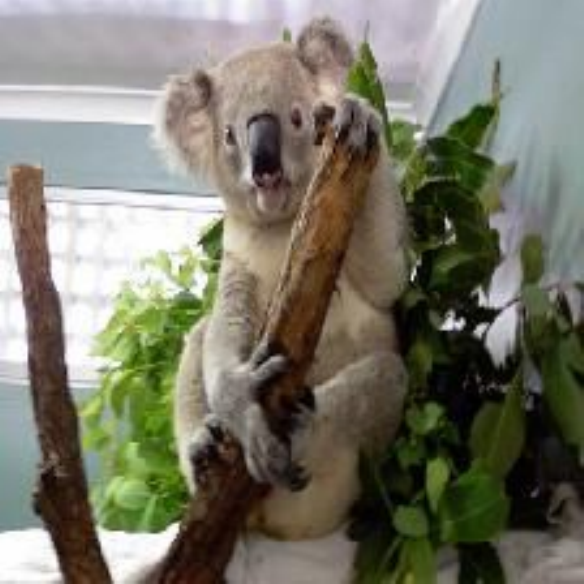}
          \vspace{12pt}
        \end{subfigure}%  
    \end{minipage}
    \hspace{-3pt}\vline\hspace{1pt}
    \begin{minipage}[c]{0.55\textwidth}
        \caption*{Adversarial Examples}
        \vspace{-8pt}
        \begin{minipage}[c]{0.242\textwidth} 
        \begin{subfigure}{\textwidth}
              \centering 
              \caption*{\small \#Q.=0}
              \vspace{-3pt}
              \includegraphics[width=\linewidth]{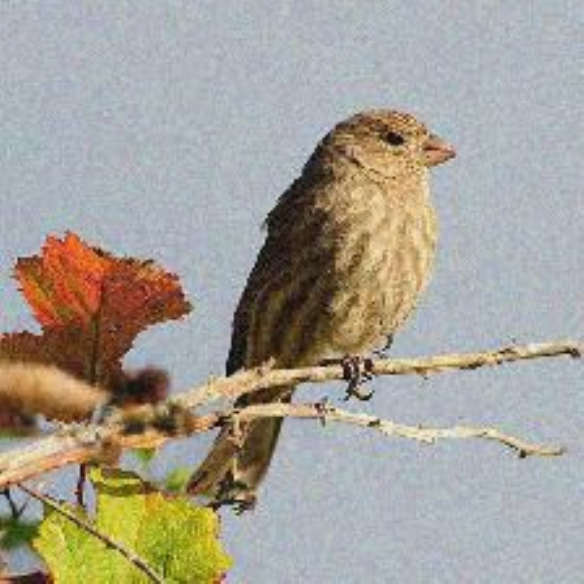}
              \vspace{-15pt}
              \caption*{RMSE=0.071}
            \end{subfigure}\\
            \vspace{2pt}
            \begin{subfigure}{\textwidth} 
              \centering 
              \includegraphics[width=\linewidth]{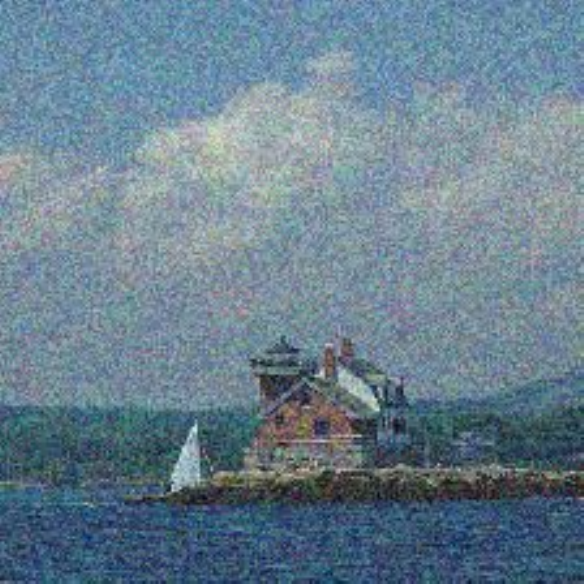}
              \vspace{-15pt}
              \caption*{RMSE=0.139}
            \end{subfigure}\\
            \vspace{2pt}
            \begin{subfigure}{\textwidth} 
              \centering 
              \includegraphics[width=\linewidth]{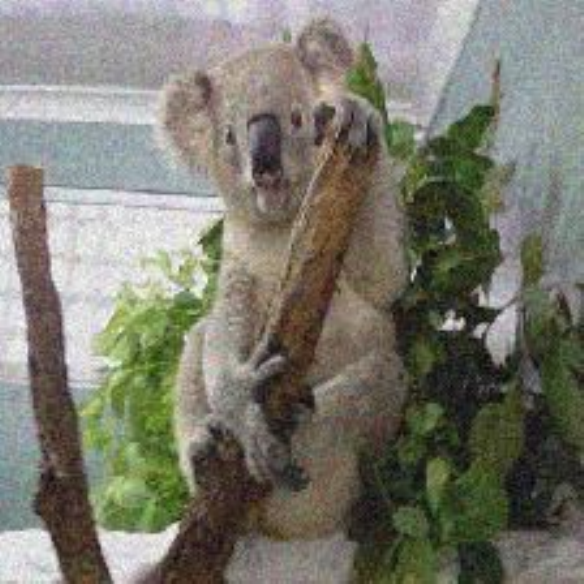}
              \vspace{-15pt}
              \caption*{RMSE=0.075}
            \end{subfigure}%  
        \end{minipage}
        \hspace{-4pt}
        \begin{minipage}[c]{0.242\textwidth} 
            \begin{subfigure}{\textwidth}
              \centering 
              \caption*{\small \#Q.=50}
              \vspace{-3pt}
              \includegraphics[width=\linewidth]{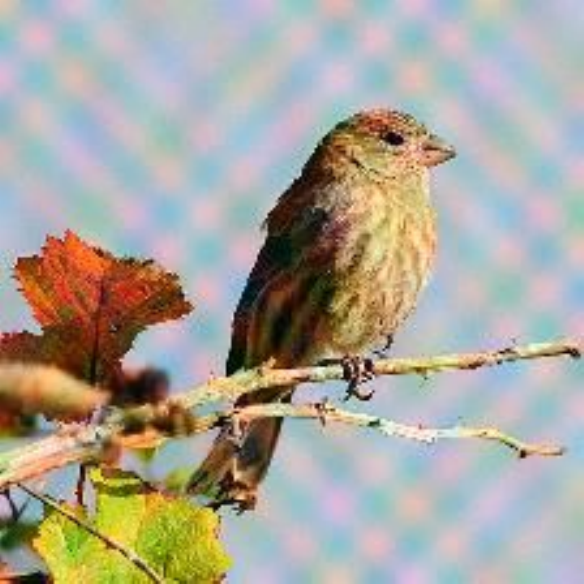}
              \vspace{-15pt}
              \caption*{RMSE=0.029}
            \end{subfigure}\\
            \vspace{2pt}
            \begin{subfigure}{\textwidth} 
              \centering 
              \includegraphics[width=\linewidth]{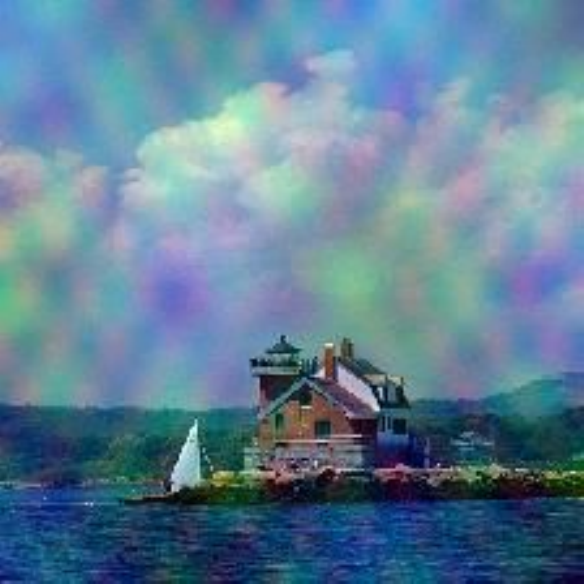}
              \vspace{-15pt}
              \caption*{RMSE=0.057}
            \end{subfigure}\\
            \vspace{2pt}
            \begin{subfigure}{\textwidth} 
              \centering 
              \includegraphics[width=\linewidth]{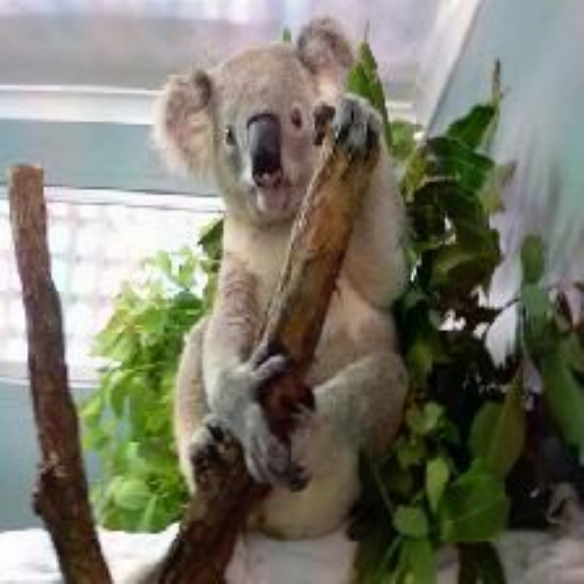}
              \vspace{-15pt}
              \caption*{RMSE=0.011}
            \end{subfigure}% 
        \end{minipage}
        \hspace{-4pt}
        \begin{minipage}[c]{0.242\textwidth} 
            \begin{subfigure}{\textwidth}
              \centering 
              \caption*{\small \#Q.=100}
              \vspace{-3pt}
              \includegraphics[width=\linewidth]{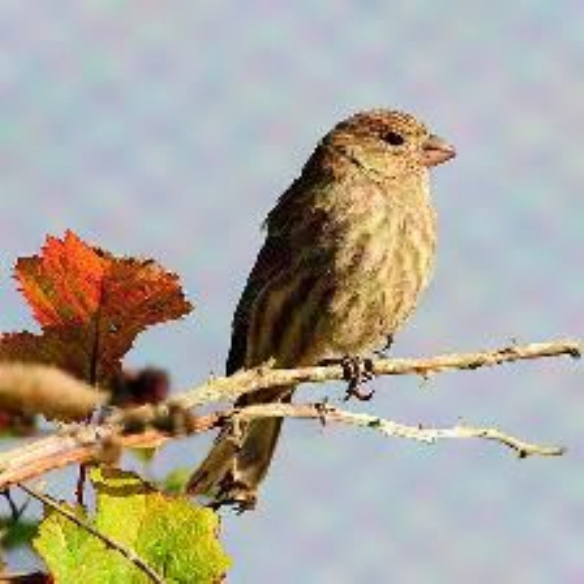}
              \vspace{-15pt}
              \caption*{RMSE=0.017}
            \end{subfigure}\\
            \vspace{2pt}
            \begin{subfigure}{\textwidth} 
              \centering 
              \includegraphics[width=\linewidth]{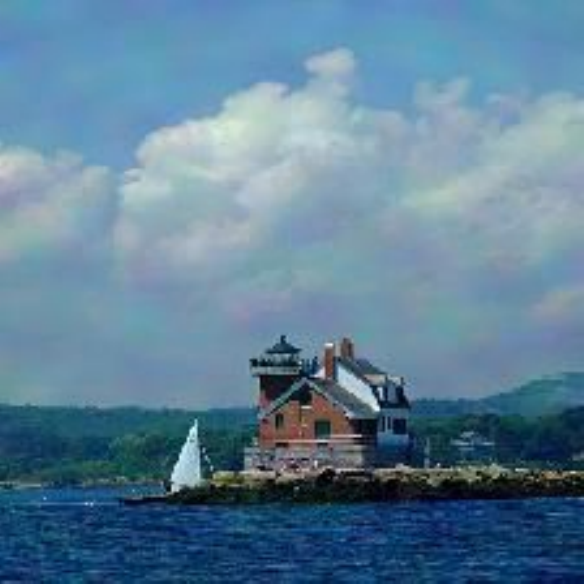}
              \vspace{-15pt}
              \caption*{RMSE=0.028}
            \end{subfigure}\\
            \vspace{2pt}
            \begin{subfigure}{\textwidth} 
              \centering 
              \includegraphics[width=\linewidth]{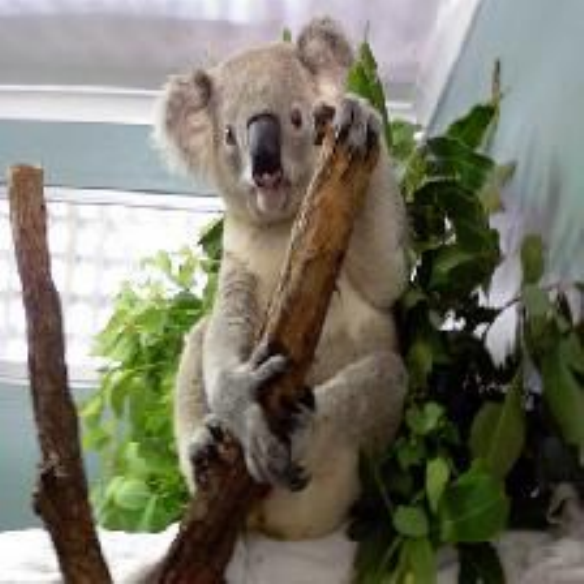}
              \vspace{-15pt}
              \caption*{RMSE=0.004}
            \end{subfigure}%  
        \end{minipage}
        \hspace{-4pt}
        \begin{minipage}[c]{0.242\textwidth} 
            \begin{subfigure}{\textwidth}
              \centering 
              \caption*{\small \#Q.=200}
              \vspace{-3pt}
              \includegraphics[width=\linewidth]{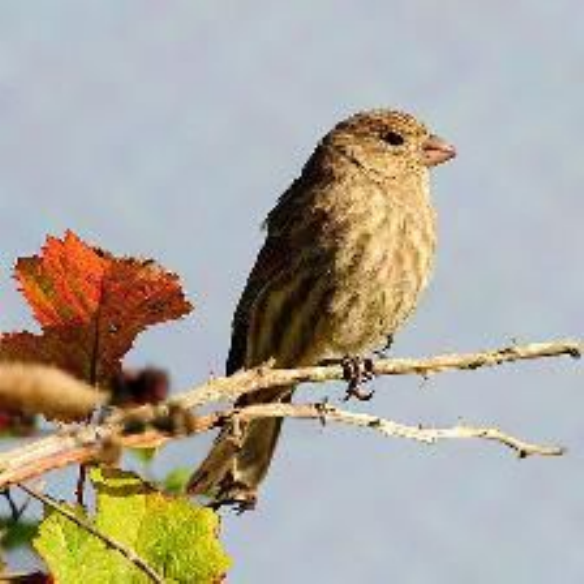}%
              \makebox[15pt][r]{\makebox[20pt]{\raisebox{16pt}{\rotatebox[origin=c]{270}{\small Red Fox}}}}
              \vspace{-15pt}
              \caption*{RMSE=0.007}
            \end{subfigure}\\
            \vspace{2pt}
            \begin{subfigure}{\textwidth} 
              \centering 
              \includegraphics[width=\linewidth]{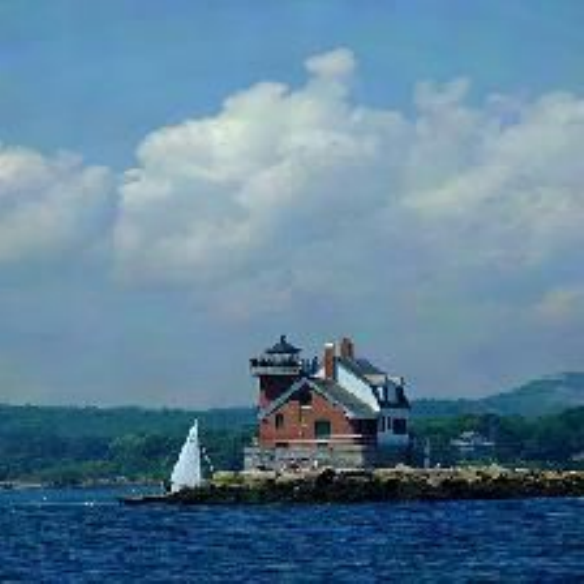}%
              \makebox[15pt][r]{\makebox[20pt]{\raisebox{16pt}{\rotatebox[origin=c]{270}{\small Motor}}}}
              \vspace{-15pt}
              \caption*{RMSE=0.008}
            \end{subfigure}\\
            \vspace{2pt}
            \begin{subfigure}{\textwidth} 
              \centering 
              \includegraphics[width=\linewidth]{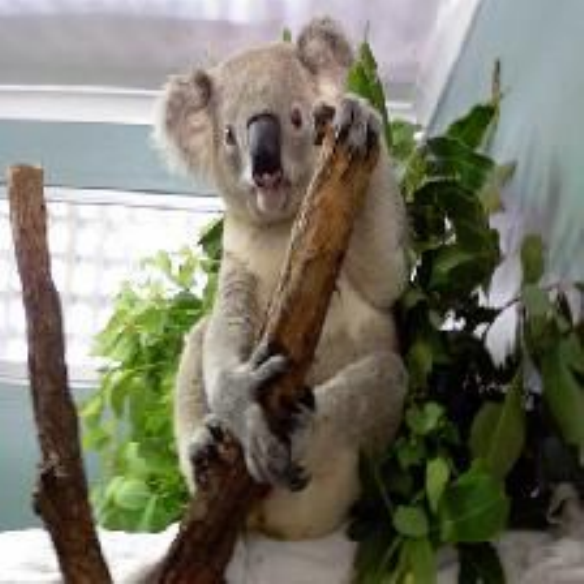}%
              \makebox[15pt][r]{\makebox[20pt]{\raisebox{18pt}{\rotatebox[origin=c]{270}{\small Fire Screen}}}}
              \vspace{-15pt}
              \caption*{RMSE=0.002}
              
            \end{subfigure}%  
        \end{minipage}
    \end{minipage}
    \vspace{-0.5em}
    \caption{The adversarial examples crafted by \name against \app. \#Q. denotes the number of queries for attack and RMSE denotes the RMSE distance between the benign sample and adversarial example. We report the correct label and the predicted label on the leftmost and rightmost columns, respectively (Zoom in for details.)}
    \label{fig:tencent_res}
    \vspace{-1.5em}
\end{figure}

The numbers of successfully attacked images are summarized in Table~\ref{tab:tencent_api}. We can observe that \name successfully generates more adversarial examples than the attack baselines within 200, 500 and 1,000 queries under various $RMSE$ thresholds. In particular, our \name can generate even more adversarial examples within 500 queries than the best attack baselines within 1,000 queries, showing the superiority of \name. We also visualize some adversarial examples generated by \name in Fig.~\ref{fig:tencent_res}. As we can see, \name can successfully generate high quality adversarial examples for various classes with few queries ($\leq 200$), validating the high applicability of \name in real-world. Especially when the number of queries is 200, the adversarial examples generated by \name are almost visually imperceptible for humans, which highlights the vulnerability of current commercial applications.

\subsection{Ablation Study}
\label{sec:exp:ablation}
% To further investigate the superior performance of \name,
In this section, we conduct a series of ablation studies on ResNet-18, namely the subspace chosen by \name, the ratio for low frequency subspace and the change rate $\gamma$ and $\lambda$ for updating the angle $\alpha$. The parameter studies on the dimension of sampled line $d$ and the bound $\tau$ for $\alpha$ are summarized in Appendix~B.

\begin{wraptable}{r}{0.435\textwidth}
    \centering
    \vspace{-2.5em}
    \caption{Ablation study on ResNet-18 for different spaces, \ie input space (TA$_\mathrm{I}$), frequency space for line sampling but input space for perturbation optimization (TA$_\mathrm{FI}$), and full frequency space without mask (TA$_\mathrm{F}$)}
    \label{tab:subspace}
    \vspace{-0.8em}
    \begin{tabular}{lcccc}
        \toprule
        RMSE & TA$_\mathrm{I}$ & TA$_\mathrm{FI}$  & TA$_\mathrm{F}$ & TA \\
        \midrule
        0.1 & 39.5 & 97.5 & 98.5 & \bf{100.0}\\
        0.05 & 17.5 & 73.0 & 85.0 & ~~\bf{94.0}\\
        0.01 & ~~3.0 & 22.5 & 25.5 & ~~\bf{51.5}\\
        \bottomrule
    \end{tabular}
    \vspace{-1.8em}
\end{wraptable}

\textbf{On the subspace chosen by \name.} Different from existing decision-based attacks, the generality of geometric property used by \name makes it possible to directly optimize the perturbation in the frequency space. To investigate the effectiveness of frequency space, we implement \name in various spaces, namely input space (TA$_\mathrm{I}$), sampling the directional line in the frequency space but optimizing the perturbation in the input space (TA$_\mathrm{FI}$) used by Surfree~\cite{maho2021surfree} and full frequency space (TA$_\mathrm{F}$). As shown in Table~\ref{tab:subspace}, due to the high-dimensional input space, TA$_\mathrm{I}$ cannot effectively explore the neighborhood of the input sample to find good perturbation and shows very poor performance. With the information from frequency space to sample the subspace, TA$_\mathrm{FI}$ exhibits much better results than TA$_\mathrm{I}$. When optimizing the perturbation in the full frequency space, TA$_\mathrm{F}$ can achieve higher attack success rate than TA$_\mathrm{FI}$, showing the benefit of frequency space. When sampling the subspace using the low frequency information, \name achieves much better performance than all the other attacks, supporting the necessity and rationality of the subspace chosen by \name. 

\begin{figure*}[t]
\begin{minipage}{.48\linewidth}
\centering
\includegraphics[width=0.95\linewidth]{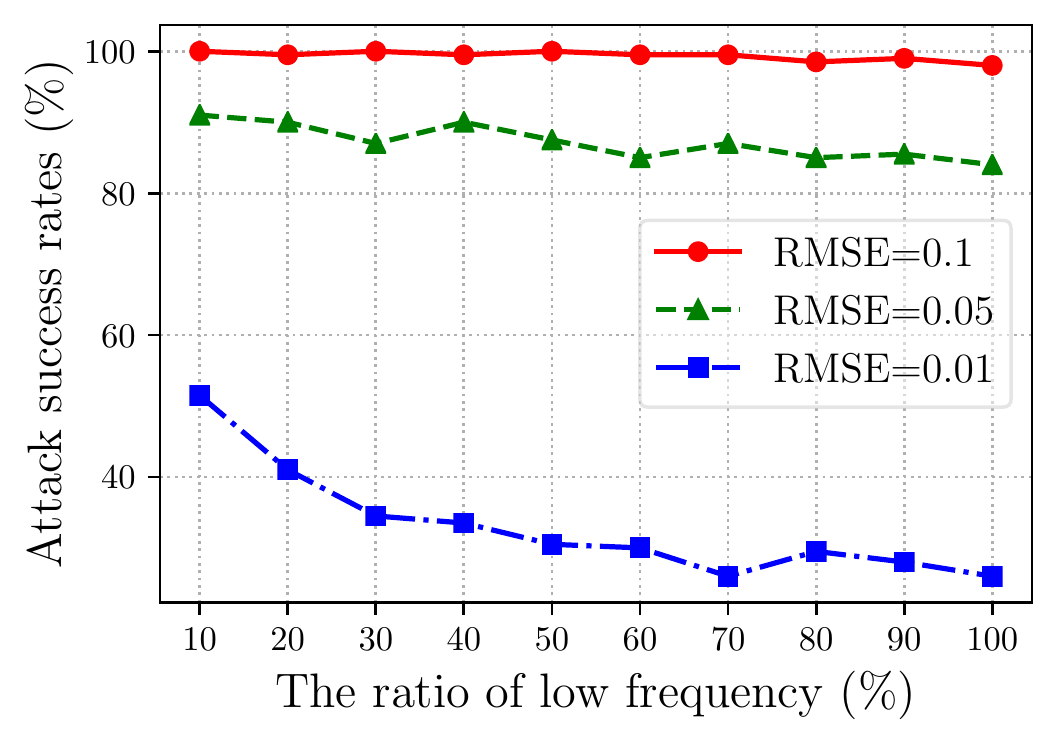}
\vspace{-1em}
\caption{ Attack success rate (\%) of \name on ResNet-18 within 1,000 queries with various ratios for the low frequency subspace under three $RMSE$ thresholds}
\label{fig:ratio_freq}
\end{minipage}
\hfill
\begin{minipage}{.48\linewidth}
\centering
\includegraphics[width=0.95\linewidth]{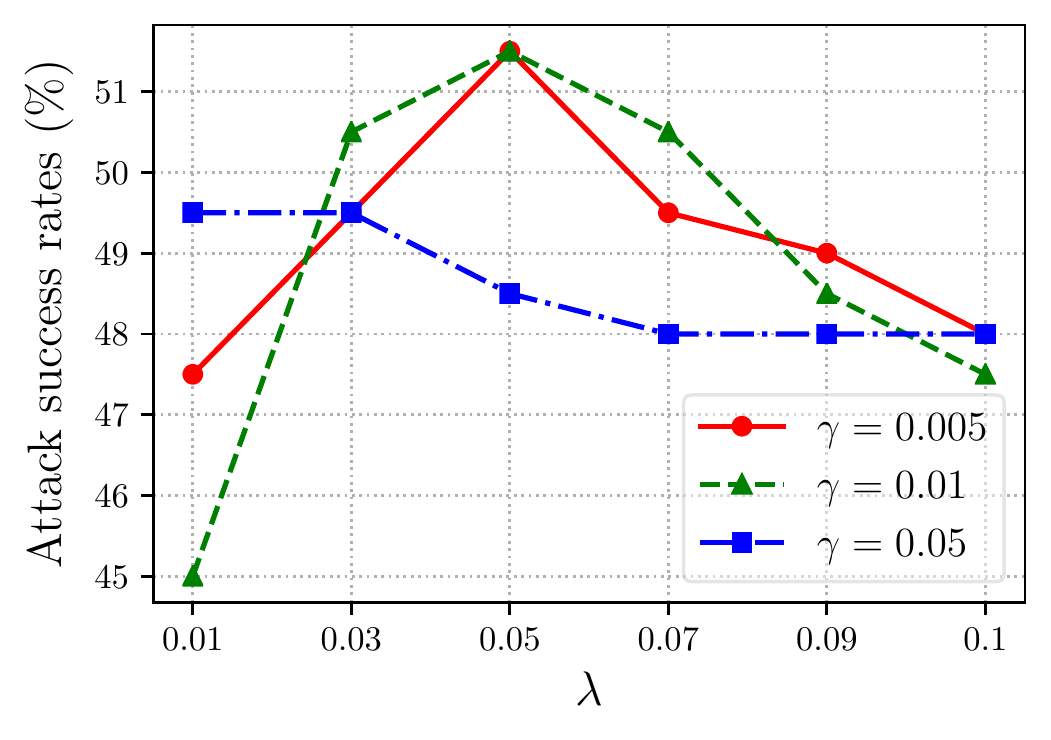}
\vspace{-1em}
\caption{Attack success rate (\%) of \name on ResNet-18 within 1,000 queries with various $\gamma$ and $\lambda$ used for updating $\alpha$ under $RMSE = 0.01$}
\label{fig:lambda}
\end{minipage}
\vspace{-2em}
\end{figure*}

\textbf{On the ratio for low frequency subspace.} The low frequency domain plays key role in improving the efficiency of \name. However, there is no criterion to identify the low frequency since it corresponds to high frequency, which is usually determined by the lower part of the frequency domain with a given ratio. Here we investigate the effect of this ratio on the attack performance of \name. As shown in Fig.~\ref{fig:ratio_freq}, the ratio has more significant influence on the attack success rate under the smaller $RMSE$ threshold. In general, increasing the ratio roughly decreases the attack performance because it makes \name focus more on the high frequency domain, containing less critical information of the image. Thus, we adopt the lower 10\% parts as the low frequency subspace for high efficiency, which also helps \name effectively reduce the dimension, making it easier for attack.

\textbf{On the change rate $\gamma$ and $\lambda$ for updating the angle $\alpha$.} As stated in Sec.~\ref{sec:method}, the angle $\alpha$ plays a key role in choosing a better candidate triangle but it is hard to find a uniformly optimal $\alpha$ for different iterations and input images. We assume that the larger angle $\alpha$ makes it harder to find a candidate triangle but leads to smaller perturbation. As in Eq.~\eqref{eq:alpha}, if we successfully find a triangle, we would increase $\alpha$ with $\gamma$. Otherwise, we would decrease $\alpha$ with $\lambda \gamma$. We investigate the impact of various $\gamma$ and $\lambda$ in Fig.~\ref{fig:lambda}. Here we only report the results for $RMSE = 0.01$ for clarity and the results for $RMSE = 0.1/0.05$ exhibit the same trend. In general, $\gamma = 0.01$ leads to better attack performance than $\gamma = 0.05/0.005$. When we increase $\lambda$ with $\gamma = 0.01$, the attack success rate increases until $\lambda = 0.05$ and then decreases. We also investigate the impact of $\tau$ which controls the bound for $\alpha$ in Eq.~\eqref{eq:alpha}, which shows stable performance within 1,000 queries but takes effect for 10,000 queries and we simply adopt $\tau = 0.1$.  In our experiments, we adopt $\gamma = 0.01$, $\lambda = 0.05$ and $\tau = 0.1$.

\subsection{Further Discussion}
\label{sec:discussion}
\begin{wrapfigure}{r}{0.48\textwidth}
    \centering
    \vspace{-4.5em}
    \includegraphics[width=0.92\linewidth]{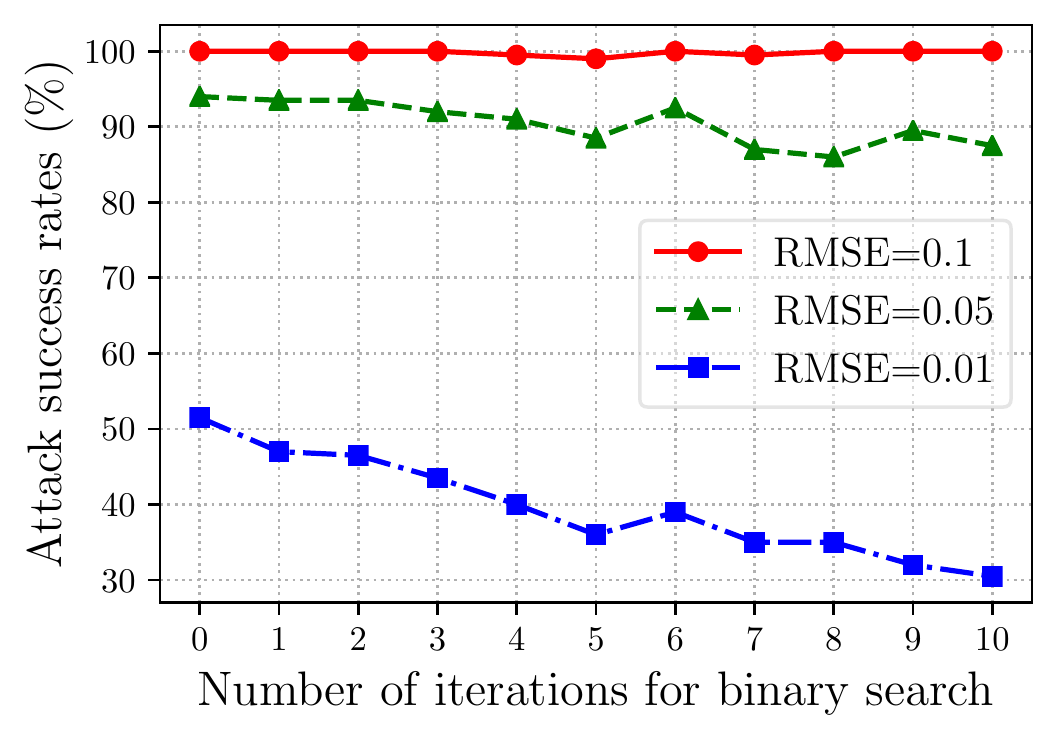}
    \vspace{-1.2em}
    \caption{Attack success rate (\%) of TA using various number of iterations for binary search ($N_{bs}$) to restrict the adversary on the decision boundary at each iteration}
    \label{fig:bs}
    \vspace{-2.2em}
\end{wrapfigure}

BoundaryAttack~\cite{brendel2017decision} adopts random walk on the decision boundary to minimize the perturbation for decision-based attack and the subsequent works often follow this setting to restrict the adversarial example on the decision boundary. We argue that such a restriction is not necessary and do not adopt it in our \name. To validate this argument, we also conduct binary search to move the adversarial example towards the decision boundary at each iteration after we find the candidate triangle to investigate the benefit of this restriction. As illustrated in Fig.~\ref{fig:bs}, when the number of iterations for binary search ($N_{bs}$) is 0, it is vanilla \name that exhibits the best attack success rate. When we increase $N_{bs}$, the binary search takes more queries in each iteration which degrades the total number of iterations under the given total number of queries. In general, the attack success rate stably decreases when increasing $N_{bs}$ especially for $RMSE = 0.01$, which means the cost (\ie, queries) for binary search to restrict the adversarial example on the decision boundary is not worthy. Such restriction might also be not reliable and rational for most decision-based attacks, especially for geometry-inspired attacks. We hope this would inspire more attention to discuss the necessity of restricting the adversarial examples on the decision boundary and shed new light on the design of more powerful decision-based attacks.

\vspace{-0.1em}
\section{Conclusion}
\vspace{-0.1em}
In this work, we found that the benign sample, current and next adversarial examples can naturally construct a triangle in a subspace at each iteration for any iterative attacks. Based on this observation, we proposed a novel decision-based attack, called Triangle Attack (\name), which utilizes the geometric information that the longer side is opposite the larger angle in any triangle. Specifically, at each iteration, \name randomly samples a directional line across the benign sample to determine a subspace, in which \name iteratively searches a candidate triangle to minimize the adversarial perturbation. With the generality of geometric property, \name directly optimizes the adversarial perturbation in the low frequency space generated by DCT with much lower dimensions than the input space, and significantly improves the query efficiency. Extensive experiments demonstrate that \name achieves a much higher attack success rate within 1,000 queries and needs much less queries to achieve the same attack success rate. The practical applicability on \app also validates the superiority of \name. 
% We hope that \name can shed light on improving the query efficiency via geometric information and other effective dimensionality reduction methods.

\vspace{-0.5em}
\subsubsection{Acknowledgement} This work is supported by National Natural Science Foundation (62076105), Intertational Coorperation Foundation of Hubei Province, China(2021EHB011) and Tencent Rhino Bird Elite Talent Training Program.

{\small
\bibliographystyle{ieee_fullname}
\bibliography{egbib}

\begin{thebibliography}{10}\itemsep=-1pt

\bibitem{ahmed1974discrete}
Nasir Ahmed, T\_ Natarajan, and Kamisetty~R Rao.
\newblock Discrete cosine transform.
\newblock {\em IEEE Transactions on Computers}, 1974.

\bibitem{al2019sign}
Abdullah Al{-}Dujaili and Una{-}May O'Reilly.
\newblock Sign bits are all you need for black-box attacks.
\newblock In {\em International Conference on Learning Representations}, 2020.

\bibitem{athalye2018obfuscated}
Anish Athalye, Nicholas Carlini, and David~A. Wagner.
\newblock Obfuscated gradients give a false sense of security: Circumventing
  defenses to adversarial examples.
\newblock In {\em International Conference on Machine Learning}, 2018.

\bibitem{bojarski2016end}
Mariusz Bojarski, Davide Del~Testa, Daniel Dworakowski, Bernhard Firner, and
  Beat Flepp~et al.
\newblock End to end learning for self-driving cars.
\newblock {\em arXiv preprint arXiv:1604.07316}, 2016.

\bibitem{brendel2017decision}
Wieland Brendel, Jonas Rauber, and Matthias Bethge.
\newblock Decision-based adversarial attacks: Reliable attacks against
  black-box machine learning models.
\newblock In {\em International Conference on Learning Representations}, 2018.

\bibitem{carlini2017towards}
Nicholas Carlini and David Wagner.
\newblock Towards evaluating the robustness of neural networks.
\newblock In {\em IEEE Symposium on Security and Privacy}, 2017.

\bibitem{chen2015deepdriving}
Chenyi Chen, Ari Seff, Alain Kornhauser, and Jianxiong Xiao.
\newblock Deepdriving: Learning affordance for direct perception in autonomous
  driving.
\newblock In {\em International Conference on Computer Vision}, pages
  2722--2730, 2015.

\bibitem{chen2020hopskipjumpattack}
Jianbo Chen, Michael~I Jordan, and Martin~J Wainwright.
\newblock Hopskipjumpattack: {A} query-efficient decision-based attack.
\newblock In {\em IEEE Symposium on Security and Privacy}, 2020.

\bibitem{chen2017zoo}
Pin-Yu Chen, Huan Zhang, Yash Sharma, Jinfeng Yi, and Cho-Jui Hsieh.
\newblock Zoo: Zeroth order optimization based black-box attacks to deep neural
  networks without training substitute models.
\newblock In {\em ACM Workshop on Artificial Intelligence and Security}, 2017.

\bibitem{chen2020boosting}
Weilun Chen, Zhaoxiang Zhang, Xiaolin Hu, and Baoyuan Wu.
\newblock Boosting decision-based black-box adversarial attacks with random
  sign flip.
\newblock In {\em European Conference on Computer Vision}, 2020.

\bibitem{cheng2018query}
Minhao Cheng, Thong Le, Pin{-}Yu Chen, Huan Zhang, Jinfeng Yi, and Cho{-}Jui
  Hsieh.
\newblock Query-efficient hard-label black-box attack: An optimization-based
  approach.
\newblock In {\em International Conference on Learning Representations}, 2019.

\bibitem{cheng2019sign}
Minhao Cheng, Simranjit Singh, Patrick~H. Chen, Pin{-}Yu Chen, Sijia Liu, and
  Cho{-}Jui Hsieh.
\newblock Sign-{OPT}: {A} query-efficient hard-label adversarial attack.
\newblock In {\em International Conference on Learning Representations}, 2020.

\bibitem{cohen2019certified}
Jeremy~M. Cohen, Elan Rosenfeld, and J.~Zico Kolter.
\newblock Certified adversarial robustness via randomized smoothing.
\newblock In {\em International Conference on Machine Learning}, 2019.

\bibitem{croce2020reliable}
Francesco Croce and Matthias Hein.
\newblock Reliable evaluation of adversarial robustness with an ensemble of
  diverse parameter-free attacks.
\newblock In {\em International Conference on Machine Learning}, 2020.

\bibitem{deng2019mutual}
Zhongying Deng, Xiaojiang Peng, Zhifeng Li, and Yu Qiao.
\newblock Mutual component convolutional neural networks for heterogeneous face
  recognition.
\newblock {\em IEEE Transactions on Image Processing}, 28(6):3102--3114, 2019.

\bibitem{dong2020benchmarking}
Yinpeng Dong, Qi{-}An Fu, Xiao Yang, Tianyu Pang, Hang Su, Zihao Xiao, and Jun
  Zhu.
\newblock Benchmarking adversarial robustness on image classification.
\newblock In {\em Conference on Computer Vision and Pattern Recognition}, 2020.

\bibitem{dong2018boosting}
Yinpeng Dong, Fangzhou Liao, Tianyu Pang, Hang Su, Jun Zhu, Xiaolin Hu, and
  Jianguo Li.
\newblock Boosting adversarial attacks with momentum.
\newblock In {\em Conference on Computer Vision and Pattern Recognition}, 2018.

\bibitem{du2019query}
Jiawei Du, Hu Zhang, Joey~Tianyi Zhou, Yi Yang, and Jiashi Feng.
\newblock Query-efficient meta attack to deep neural networks.
\newblock In {\em International Conference on Learning Representations}, 2020.

\bibitem{eykholt2018robust}
Kevin Eykholt, Ivan Evtimov, Earlence Fernandes, Bo Li, Amir Rahmati, Chaowei
  Xiao, Atul Prakash, Tadayoshi Kohno, and Dawn Song.
\newblock Robust physical-world attacks on deep learning visual classification.
\newblock In {\em Conference on Computer Vision and Pattern Recognition}, 2018.

\bibitem{gong2013multi}
Dihong Gong, Zhifeng Li, Jianzhuang Liu, and Yu Qiao.
\newblock Multi-feature canonical correlation analysis for face photo-sketch
  image retrieval.
\newblock In {\em Proceedings of the 21st ACM International Conference on
  Multimedia}, pages 617--620, 2013.

\bibitem{goodfellow2014explaining}
Ian~J. Goodfellow, Jonathon Shlens, and Christian Szegedy.
\newblock Explaining and harnessing adversarial examples.
\newblock In {\em International Conference on Learning Representations}, 2015.

\bibitem{guo2018low}
Chuan Guo, Jared~S Frank, and Kilian~Q Weinberger.
\newblock Low frequency adversarial perturbation.
\newblock {\em Uncertainty in Artificial Intelligence}, 2019.

\bibitem{guo2017countering}
Chuan Guo, Mayank Rana, Moustapha Ciss{\'{e}}, and Laurens van~der Maaten.
\newblock Countering adversarial images using input transformations.
\newblock In {\em International Conference on Learning Representations}, 2018.

\bibitem{he2016deep}
Kaiming He, Xiangyu Zhang, Shaoqing Ren, and Jian Sun.
\newblock Deep residual learning for image recognition.
\newblock In {\em Conference on Computer Vision and Pattern Recognition}, 2016.

\bibitem{huang2017densely}
Gao Huang, Zhuang Liu, Laurens van~der Maaten, and Kilian~Q. Weinberger.
\newblock Densely connected convolutional networks.
\newblock In {\em Conference on Computer Vision and Pattern Recognition}, 2017.

\bibitem{ilyas2018black}
Andrew Ilyas, Logan Engstrom, Anish Athalye, and Jessy Lin.
\newblock Black-box adversarial attacks with limited queries and information.
\newblock In {\em International Conference on Machine Learning}, 2018.

\bibitem{krizhevsky2012imagenet}
Alex Krizhevsky, Ilya Sutskever, and Geoffrey~E. Hinton.
\newblock {ImageNet} classification with deep convolutional neural networks.
\newblock In {\em Advances in Neural Information Processing Systems}, 2012.

\bibitem{kurakin2017IFGSM}
Alexey Kurakin, Ian Goodfellow, and Samy Bengio.
\newblock Adversarial examples in the physical world.
\newblock In {\em International Conference on Learning Representations,
  Workshop}, 2017.

\bibitem{li2020qeba}
Huichen Li, Xiaojun Xu, Xiaolu Zhang, Shuang Yang, and Bo Li.
\newblock {QEBA:} query-efficient boundary-based blackbox attack.
\newblock In {\em Conference on Computer Vision and Pattern Recognition}, 2020.

\bibitem{li2014common}
Zhifeng Li, Dihong Gong, Yu Qiao, and Dacheng Tao.
\newblock Common feature discriminant analysis for matching infrared face
  images to optical face images.
\newblock {\em IEEE Transactions on Image Processing}, 23(6):2436--2445, 2014.

\bibitem{liang2022parallel}
Siyuan Liang, Baoyuan Wu, Yanbo Fan, Xingxing Wei, and Xiaochun Cao.
\newblock Parallel rectangle flip attack: A query-based black-box attack
  against object detection.
\newblock {\em arXiv preprint arXiv:2201.08970}, 2022.

\bibitem{liu2016delving}
Yanpei Liu, Xinyun Chen, Chang Liu, and Dawn Song.
\newblock Delving into transferable adversarial examples and black-box attacks.
\newblock In {\em International Conference on Learning Representations}, 2017.

\bibitem{liu2019geometry}
Yujia Liu, Seyed-Mohsen Moosavi-Dezfooli, and Pascal Frossard.
\newblock A geometry-inspired decision-based attack.
\newblock In {\em International Conference on Computer Vision}, pages
  4890--4898, 2019.

\bibitem{madry2017towards}
Aleksander Madry, Aleksandar Makelov, Ludwig Schmidt, Dimitris Tsipras, and
  Adrian Vladu.
\newblock Towards deep learning models resistant to adversarial attacks.
\newblock In {\em International Conference on Learning Representations}, 2018.

\bibitem{maho2021surfree}
Thibault Maho, Teddy Furon, and Erwan Le~Merrer.
\newblock {SurFree}: {A} fast surrogate-free black-box attack.
\newblock In {\em Conference on Computer Vision and Pattern Recognition}, 2021.

\bibitem{moosavi2016deepfool}
Seyed{-}Mohsen Moosavi{-}Dezfooli, Alhussein Fawzi, and Pascal Frossard.
\newblock {DeepFool}: {A} simple and accurate method to fool deep neural
  networks.
\newblock In {\em Conference on Computer Vision and Pattern Recognition}, 2016.

\bibitem{qiu2021end2end}
Haibo Qiu, Dihong Gong, Zhifeng Li, Wei Liu, and Dacheng Tao.
\newblock End2end occluded face recognition by masking corrupted features.
\newblock {\em IEEE Transactions on Pattern Analysis and Machine Intelligence},
  2021.

\bibitem{rahmati2020geoda}
Ali Rahmati, Seyed-Mohsen Moosavi-Dezfooli, Pascal Frossard, and Huaiyu Dai.
\newblock {GeoDA:} a geometric framework for black-box adversarial attacks.
\newblock In {\em Computer Vision and Pattern Recognition}, pages 8446--8455,
  2020.

\bibitem{russakovsky2015imagenet}
Olga Russakovsky, Jia Deng, Hao Su, Jonathan Krause, Sanjeev Satheesh, Sean Ma,
  Zhiheng Huang, Andrej Karpathy, Aditya Khosla, Michael Bernstein, et~al.
\newblock {ImageNet} large scale visual recognition challenge.
\newblock In {\em International booktitle of computer vision}, 2015.

\bibitem{sallab2017deep}
Ahmad~EL Sallab, Mohammed Abdou, Etienne Perot, and Senthil Yogamani.
\newblock Deep reinforcement learning framework for autonomous driving.
\newblock {\em Electronic Imaging}, 2017(19):70--76, 2017.

\bibitem{shafahi2019adversarial}
Ali Shafahi, Mahyar Najibi, Amin Ghiasi, Zheng Xu, John~P. Dickerson, Christoph
  Studer, Larry~S. Davis, Gavin Taylor, and Tom Goldstein.
\newblock Adversarial training for free!
\newblock In {\em Advances in Neural Information Processing Systems}, 2019.

\bibitem{sharif2016accessorize}
Mahmood Sharif, Sruti Bhagavatula, Lujo Bauer, and Michael~K Reiter.
\newblock Accessorize to a crime: Real and stealthy attacks on state-of-the-art
  face recognition.
\newblock In {\em ACM SIGSAC conference on Computer and Communications
  Security}, 2016.

\bibitem{shukla2021simple}
Satya~Narayan Shukla, Anit~Kumar Sahu, Devin Willmott, and Zico Kolter.
\newblock Simple and efficient hard label black-box adversarial attacks in low
  query budget regimes.
\newblock In {\em Proceedings of the 27th ACM SIGKDD Conference on Knowledge
  Discovery \& Data Mining}, pages 1461--1469, 2021.

\bibitem{simonyan2014very}
Karen Simonyan and Andrew Zisserman.
\newblock Very deep convolutional networks for large-scale image recognition.
\newblock In {\em International Conference on Learning Representations}, 2015.

\bibitem{szegedy2016rethinking}
Christian Szegedy, Vincent Vanhoucke, Sergey Ioffe, Jonathon Shlens, and
  Zbigniew Wojna.
\newblock Rethinking the inception architecture for computer vision.
\newblock In {\em Conference on Computer Vision and Pattern Recognition}, 2016.

\bibitem{szegedy2016inceptionv3}
Christian Szegedy, Vincent Vanhoucke, Sergey Ioffe, Jonathon Shlens, and
  Zbigniew Wojna.
\newblock Rethinking the inception architecture for computer vision.
\newblock In {\em Conference on Computer Vision and Pattern Recognition}, 2016.

\bibitem{szegedy2013intriguing}
Christian Szegedy, Wojciech Zaremba, Ilya Sutskever, Joan Bruna, Dumitru Erhan,
  Ian~J. Goodfellow, and Rob Fergus.
\newblock Intriguing properties of neural networks.
\newblock In {\em International Conference on Learning Representations}, 2014.

\bibitem{tang2004video}
Xiaoou Tang and Zhifeng Li.
\newblock Video based face recognition using multiple classifiers.
\newblock In {\em IEEE International Conference on Automatic Face and Gesture
  Recognition}, pages 345--349. IEEE, 2004.

\bibitem{tu2019autozoom}
Chun-Chen Tu, Paishun Ting, Pin-Yu Chen, Sijia Liu, Huan Zhang, Jinfeng Yi,
  Cho-Jui Hsieh, and Shin-Ming Cheng.
\newblock {Autozoom}: Autoencoder-based zeroth order optimization method for
  attacking black-box neural networks.
\newblock In {\em AAAI Conference on Artificial Intelligence}, 2019.

\bibitem{wang2018cosface}
Hao Wang, Yitong Wang, Zheng Zhou, Xing Ji, Dihong Gong, Jingchao Zhou, Zhifeng
  Li, and Wei Liu.
\newblock Cosface: Large margin cosine loss for deep face recognition.
\newblock In {\em Conference on Computer Vision and Pattern Recognition}, 2018.

\bibitem{wang2021Enhancing}
Xiaosen Wang and Kun He.
\newblock Enhancing the transferability of adversarial attacks through variance
  tuning.
\newblock In {\em Conference on Computer Vision and Pattern Recognition}, 2021.

\bibitem{wang2021admix}
Xiaosen Wang, Xuanran He, Jingdong Wang, and Kun He.
\newblock Admix: Enhancing the transferability of adversarial attacks.
\newblock In {\em International Conference on Computer Vision}, 2021.

\bibitem{wang2021Natural}
Xiaosen Wang, Hao Jin, Yichen Yang, and Kun He.
\newblock Natural language adversarial defense through synonym encoding.
\newblock {\em Conference on Uncertainty in Artificial Intelligence}, 2021.

\bibitem{wang2021boosting}
Xiaosen Wang, Jiadong Lin, Han Hu, Jingdong Wang, and Kun He.
\newblock Boosting adversarial transferability through enhanced momentum.
\newblock In {\em British Machine Vision Conference}, 2021.

\bibitem{wei2018transferable}
Xingxing Wei, Siyuan Liang, Ning Chen, and Xiaochun Cao.
\newblock Transferable adversarial attacks for image and video object
  detection.
\newblock {\em arXiv preprint arXiv:1811.12641}, 2018.

\bibitem{wen2016discriminative}
Yandong Wen, Kaipeng Zhang, Zhifeng Li, and Yu Qiao.
\newblock A discriminative feature learning approach for deep face recognition.
\newblock In {\em European Conference on Computer Vision}, 2016.

\bibitem{wong2020fast}
Eric Wong, Leslie Rice, and J.~Zico Kolter.
\newblock Fast is better than free: Revisiting adversarial training.
\newblock In {\em International Conference on Learning Representations}, 2020.

\bibitem{wu2021attacking}
Boxi Wu, Heng Pan, Li Shen, Jindong Gu, Shuai Zhao, Zhifeng Li, Deng Cai,
  Xiaofei He, and Wei Liu.
\newblock Attacking adversarial attacks as a defense.
\newblock {\em arXiv preprint arXiv:2106.04938}, 2021.

\bibitem{wu2021improving}
Weibin Wu, Yuxin Su, Michael~R Lyu, and Irwin King.
\newblock Improving the transferability of adversarial samples with adversarial
  transformations.
\newblock In {\em Conference on Computer Vision and Pattern Recognition}, 2021.

\bibitem{xie2019improving}
Cihang Xie, Zhishuai Zhang, Yuyin Zhou, Song Bai, Jianyu Wang, Zhou Ren, and
  Alan~L. Yuille.
\newblock Improving transferability of adversarial examples with input
  diversity.
\newblock In {\em Conference on Computer Vision and Pattern Recognition}, 2019.

\bibitem{xu2017end}
Huazhe Xu, Yang Gao, Fisher Yu, and Trevor Darrell.
\newblock End-to-end learning of driving models from large-scale video
  datasets.
\newblock In {\em Conference on Computer Vision and Pattern Recognition}, pages
  2174--2182, 2017.

\bibitem{yang2021larnet}
Xiaolong Yang, Xiaohong Jia, Dihong Gong, Dong-Ming Yan, Zhifeng Li, and Wei
  Liu.
\newblock Larnet: Lie algebra residual network for face recognition.
\newblock In {\em International Conference on Machine Learning}, pages
  11738--11750. PMLR, 2021.

\bibitem{yao2019trust}
Zhewei Yao, Amir Gholami, Peng Xu, Kurt Keutzer, and Michael~W. Mahoney.
\newblock Trust region based adversarial attack on neural networks.
\newblock In {\em Conference on Computer Vision and Pattern Recognition}, 2019.

\bibitem{zhang2019theoretically}
Hongyang Zhang, Yaodong Yu, Jiantao Jiao, Eric~P. Xing, Laurent~El Ghaoui, and
  Michael~I. Jordan.
\newblock Theoretically principled trade-off between robustness and accuracy.
\newblock In {\em International Conference on Machine Learning}, 2019.

\bibitem{zhao2020towards}
Pu Zhao, Pin{-}Yu Chen, Siyue Wang, and Xue Lin.
\newblock Towards query-efficient black-box adversary with zeroth-order natural
  gradient descent.
\newblock In {\em AAAI Conference on Artificial Intelligence}, 2020.

\end{thebibliography}
}
\clearpage
\appendix
\section*{Appendix}
Here we further provide the evaluations on 1,000 images for \name as well as Surfree, and the parameter studies on the dimension of directional line $d$ as well as the bound $\tau$ for angle $\alpha$.
\section{Evaluations on More Images}
\label{app:moreimages}
\begin{wraptable}{r}{0.4\textwidth}
    \centering
    \vspace{-2em}
    \caption{Attack success rate (\%) on VGG-16 with 1000 images under different RMSE thresholds. The maximum number of queries is set to 1,000. We highlight the highest attack success rate in \textbf{bold}}
    \label{tab:1000_result}
    \vspace{-0.8em}
    \begin{tabular}{lccc}
        \toprule
        RMSE & 0.1 & 0.05 & 0.01\\
        \midrule
        Surfree  & 98.4 & 90.2 & 36.5 \\
      \name(\bf{Ours})  &\bf{99.6} & \bf{93.9} & \bf{39.7} \clearrow\\
        \bottomrule
    \end{tabular}
    \vspace{-1em}
\end{wraptable}
Due to the high computation cost, most attacks adopt hundreds of images for evaluation (\eg, HSJA~\cite{chen2020hopskipjumpattack} (100), QEBA~\cite{li2020qeba} (50), GeoDA~\cite{rahmati2020geoda} (350), Surfree~\cite{maho2021surfree} (200)). We follow Surfree~\cite{maho2021surfree} with 200 images in the experiments. To better validate the effectiveness of \name, we further compare TA with Surfree~\cite{maho2021surfree} on VGG-16~\cite{simonyan2014very} using 1,000 images. As shown in Table~\ref{tab:1000_result}, TA consistently performs better than Surfree~\cite{maho2021surfree} under three RMSE thresholds, showing the superiority of TA.

\section{Further Parameter Studies}
\label{app:parastudy}

In this section, we further provide parameter studies for the dimension of the directional line $d$ and the bound $\tau$ for angle $\alpha$.
\begin{figure*}[b]
\begin{minipage}{.48\linewidth}
\centering
\includegraphics[width=\linewidth]{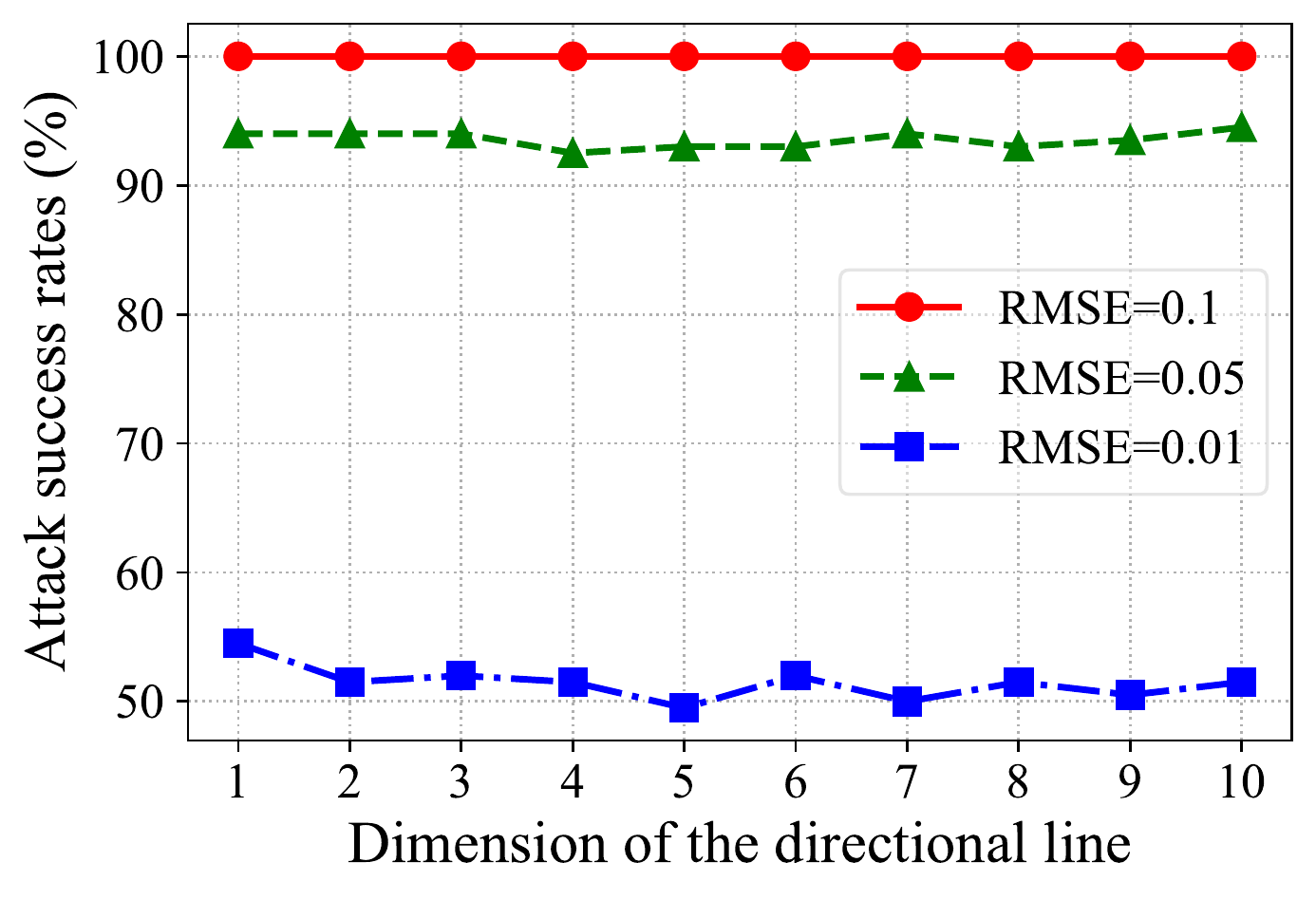}
\vspace{-2em}
\caption{ Attack success rate (\%) of \name on ResNet-18 within 1,000 queries with various dimensions of the directional line under three $RMSE$ thresholds}
\label{fig:dimension}
\end{minipage}
\hfill
\begin{minipage}{.48\linewidth}
\centering
\includegraphics[width=\linewidth]{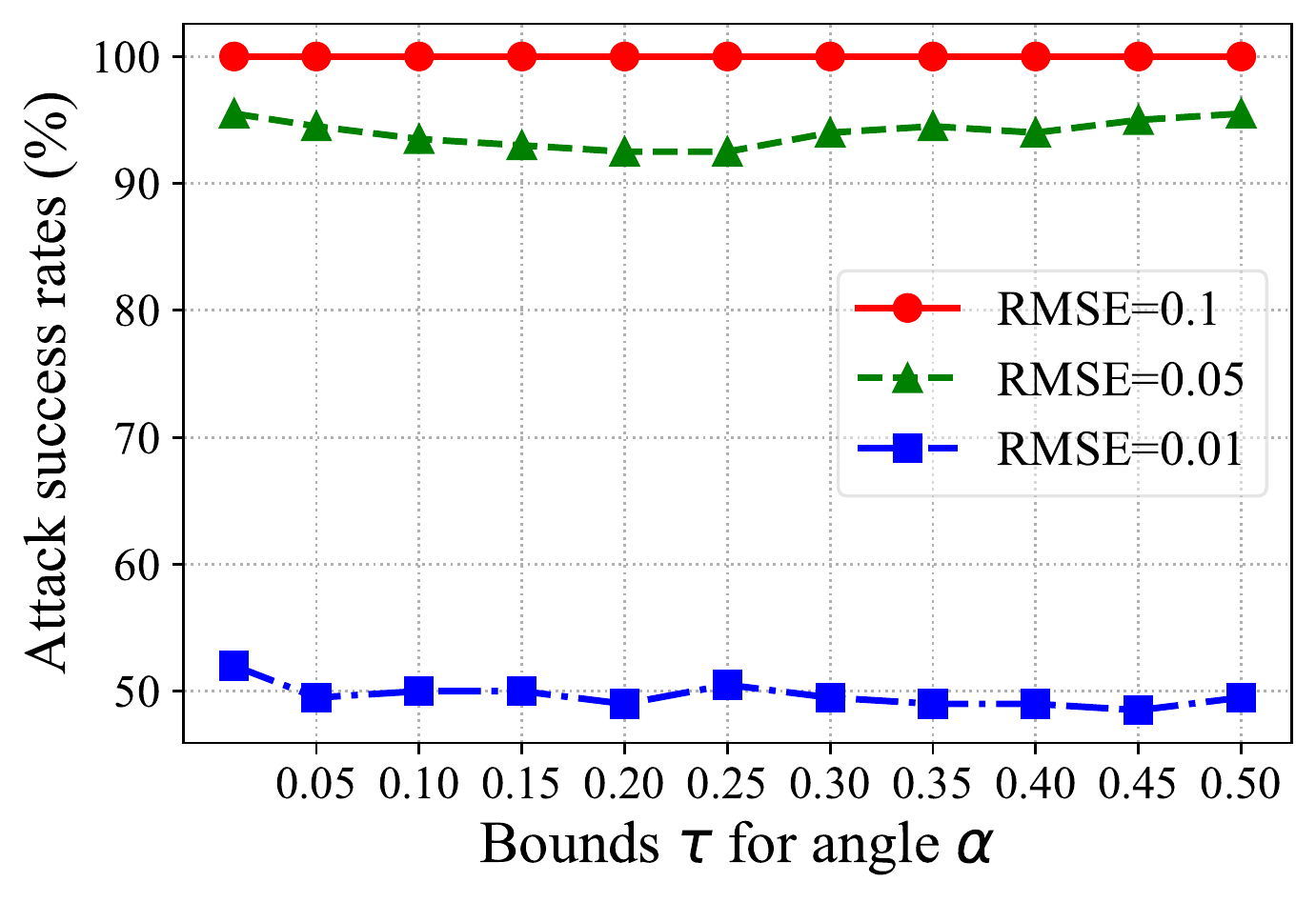}
\vspace{-2em}
\caption{Attack success rate (\%) of \name on ResNet-18 within 1,000 queries with various bounds $\tau$ for angle $\alpha$ under three $RMSE$ thresholds}
\label{fig:bound}
\end{minipage}
% \vspace{-1em}
\end{figure*}

\textbf{On the dimension of the directional line $d$}. A small dimension of line $d$ helps us sample diverse low-dimensional space in each iteration to boost the attack performance. To determine a good value for $d$, we conduct parameter studies by varying $d$ from $1$ to $10$. As shown in Fig.~\ref{fig:dimension}, with $d$ continuing to increase, the attack success rate continues to decrease, which is most obvious under the setting of $RMSE = 0.01$. Hence, we adopt $d=3$ in our experiments.

\textbf{On the bound $\tau$ for the angle $\alpha$}. A small bound $\tau$ for $\alpha$ makes the learning strategy ineffective while a large bound might result in inaccurate estimation, which degrades the performance. We also conduct parameter studies for $\tau$. As shown in Fig.~\ref{fig:bound}, a larger $\tau$ will lead to lower attack success rate, which is also more obvious when $RMSE = 0.01$. Hence, we adopt $\tau = 0.1$ in our experiments.

\end{document}